\documentclass[10pt,twocolumn,letterpaper]{article}

\usepackage{cvpr}              %

\definecolor{cvprblue}{rgb}{0.21,0.49,0.74}
\usepackage[pagebackref,breaklinks,colorlinks,allcolors=cvprblue]{hyperref}

\def\paperID{60} %
\def\confName{CVPR}
\def\confYear{2025}

\usepackage{amsmath,amsfonts,bm}

\def\eqref#1{equation~\ref{#1}}

\def\1{\bm{1}}

\DeclareMathAlphabet{\mathsfit}{\encodingdefault}{\sfdefault}{m}{sl}
\SetMathAlphabet{\mathsfit}{bold}{\encodingdefault}{\sfdefault}{bx}{n}

\newcommand{\pdata}{p_{\rm{data}}}

\newcommand{\E}{\mathbb{E}}

\newcommand{\mmm}{\texttt{[MASK]}}
\newcommand{\mm}{\texttt{[M]}}
\newcommand{\cfg}{\texttt{[C]}}

\usepackage{graphicx}
\usepackage{amsmath}
\usepackage{amssymb}
\usepackage{booktabs}
\usepackage{times}
\usepackage{microtype}
\usepackage{epsfig}
\usepackage{caption}
\usepackage{xcolor}
\usepackage{float}
\usepackage{placeins}
\usepackage{color, colortbl}
\usepackage{stfloats}
\usepackage{enumitem}
\usepackage{tabularx}
\usepackage{xstring}
\usepackage{multirow}
\usepackage{xspace}
\usepackage{url}
\usepackage{subcaption}
\usepackage{arydshln}
\usepackage{xcolor}
\usepackage{bbm}
\usepackage[hang,flushmargin]{footmisc}
\usepackage{pifont}%
\usepackage{hyperref}
\usepackage{wrapfig}
\usepackage{lipsum}
\usepackage{comment}
\usepackage{algorithm}
\usepackage{algorithmic}
\usepackage{listings}
\usepackage{makecell}

\usepackage[capitalize]{cleveref}
\crefname{section}{Sec.}{Secs.}
\Crefname{section}{Section}{Sections}
\Crefname{table}{Table}{Tables}
\crefname{table}{Tab.}{Tabs.}

\usepackage{wrapfig}

\newcommand{\noisemarg}{p_{t|1}}
\newcommand{\denoise}{p_{1|t}}
\newcommand{\x}{x}
\newcommand{\y}{y}

\newcommand{\Dt}{\Delta t}

\title{
\texttt{[MASK]}  \texttt{[MASK]} \texttt{[MASK]}  \texttt{[MASK]} \texttt{[MASK]} 
\\
\texttt{[MASK]} \textcolor{Goldenrod}{is} \texttt{[MASK]} \textcolor{Orchid}{You} \texttt{[MASK]}
\\
\texttt{[MASK]} \textcolor{Goldenrod}{is} \textcolor{Turquoise}{All} \textcolor{Orchid}{You} \textcolor{Lavender}{Need}

}

\author{Vincent Tao Hu,  Björn Ommer \\
{ CompVis @ LMU Munich, MCML}\\
\\
\url{https://compvis.github.io/mask/}
}

\begin{document}
\maketitle

\begin{abstract}
    In generative models, two paradigms have gained attraction in various applications: next-set prediction-based Masked Generative Models and next-noise prediction-based Non-Autoregressive Models, e.g., Diffusion Models. In this work, we propose using discrete-state models to connect them and explore their scalability in the vision domain. First, we conduct a step-by-step analysis in a unified design space across two types of models including timestep-independence, noise schedule, temperature, guidance strength, etc in a scalable manner. Second, we re-cast typical discriminative tasks, e.g., image segmentation, as an unmasking process from \mmm tokens on a discrete-state model. This enables us to perform various sampling processes, including flexible conditional sampling by only training once to model the joint distribution. All aforementioned explorations lead to our framework named Discrete Interpolants, which enables us to achieve state-of-the-art or competitive performance compared to previous discrete-state based methods in various benchmarks, like ImageNet256, MS COCO, and video dataset FaceForensics. In summary, by leveraging \mmm in discrete-state models, we can bridge Masked Generative and Non-autoregressive Diffusion models, as well as generative and discriminative tasks.

\end{abstract}

\section{Introduction}

Discrete tokens~\cite{esser2021taming_vqgan,rombach2022high_latentdiffusion_ldm,yu2023magvit} have gained great attention due to their compatibility with LLMs~\cite{xie2024showo,zhou2024transfusion} and compactness~\cite{yu2024an_titok,weber2024maskbit}. Based on this, Masked Generative models~\cite{chang2022maskgit,li2023mage} like MaskGiT~\cite{chang2022maskgit} have proposed gradually unmasking tokens according to specific heuristically designed rules in the vision domain. Non-Autoregressive Models, e.g., Diffusion Models—especially continuous diffusion models~\cite{sohl2015deep,song2021scorebased_sde,ho2020denoising,hu2024latent}—have contributed significantly to the generative community due to their efficacy in score prediction~\cite{song2019generative},  conditional synthesis~\cite{hu2023self,schusterbauer2024boosting,gui2024depthfm,rombach2022high_latentdiffusion_ldm,hu2024zigma}, likelihood estimation~\cite{song2021scorebased_sde}, and image inversion~\cite{he2024dice}. As research progresses from continuous-state to discrete-state diffusion models, the training and sampling similarity between Diffusion Models and Masked Generative Models become increasingly noticeable. Yet, a comprehensive analysis of their shared design space and theoretical underpinnings in the vision domain remains conspicuously absent. %

To fill this gap, we explore a framework Discrete Interpolants that builds upon the Discrete Flow Matching~\cite{gat2024discrete,shi2024simplified}, which offers flexible noise scheduling and generalization to other methods by considering discrete-state data. While this previous work initially focused on language modeling and explored only the small-scale CIFAR10 vision dataset, we explore the framework to a large-scale realistic dataset. We investigate conventional Explicit Timestep Diffusion models, which explicitly depend on timestep, as well as more flexible Implicit Timestep Diffusion models that completely remove timestep dependence. Additionally, we validate sampling behavior with our framework following the Masked Generative Models approach. This comprehensive investigation deepens our understanding of the connection between Masked Generative Models and Diffusion Models.

On the other hand, there's a trend towards unifying discriminative and generative tasks~\cite{grathwohl2019your,li2023your,fuest2024diffusion}. In this work, we demonstrate how to recast the image segmentation task into an unmasking process of our Discrete Interpolants framework. For example, given pairs of image and its segmentation mask, by training them \textit{jointly just once} to model the joint distribution in discrete-state, we can adapt our framework to various discriminative and generative tasks such as image-conditioned semantic segmentation, segmentation mask-conditioned image generation. In summary, our contributions include:

\begin{itemize}
\item We abstract and conceptualize various schedulers from discrete flow matching theory, summarizing and generalizing our framework to incorporate different coupling and conditioning methods as special cases. We utilize the progressive generalization from Explicit Timestep Models to Implicit Timestep Models to bring a closer connection between the Diffusion Model and Masked Generative Models.
\item We provide an in-depth analysis of the unified design space between Diffusion Models and Masked Generative Models, offering valuable insights for future research. Additionally, we propose that dense-pixel prediction can be reframed as an unmasking process. Furthermore, we present a comprehensive analysis of conditional generation following joint multi-modal training on the Cityscapes dataset.
\item Most importantly, by incorporating all previous designs, we achieve state-of-the-art performance on the MS-COCO dataset and competitive results on both ImageNet 256 and the video dataset Face Forensics, compared to counterpart discrete-state models.
\end{itemize}

\section{Related Work}

\textit{Due to space constraints, we have included additional related works in the Appendix.}

\paragraph{Discrete Diffusion Models.}

Several works have demonstrated deep connections between diffusion models and auto-regressive models~\cite{gat2024discrete,ou2024absorbingdiscretediffusionsecretly_radd,sahoo2024simple,zheng2024masked,liu2024think,shi2024simplified}. While this has been mainly explored in text generation, in a scaled manner~\cite{gong2024scaling,nie2024scaling}, we aim to investigate it in the vision domain. Our work differs from most others by exploring a unified design across Masked Generative Models and Diffusion models, providing a more general masking schedule in discrete-state models.
 MaskGIT~\cite{chang2022maskgit}, MAGVIT~\cite{yu2023language_magvit2}, Phenaki~\cite{villegas2022phenaki}, and MUSE~\cite{Chang2023MuseTG} focus on masked generative modeling for generation in a random order (predicting groups of tokens instead of individual tokens), we directly deploy discrete diffusion models on discrete tokens and bring the connection to them through the property of timestep-independence, and provide an in-depth analysis about different aspects of discrete diffusion. VQ-Diffusion~\cite{vqdiffusion} is a discrete-state model specifically designed for vision generation. Unlike it, we further generalize to connect between Masked Generative Model and Diffusion models by utilizing the discrete-state under our Discrete Interpolants framework.

\paragraph{Connection between  Diffusion, and Masked Generative Models.} Most Masked Generative Models~\cite{chang2022maskgit,li2023mage} use heuristically designed, greedy sampling rules based on metrics like purity~\cite{tang2022improved_vqdiffusion} or confidence~\cite{chang2022maskgit}. However, these approaches have been shown to cause over-sampling issues~\cite{gat2024discrete}. While our framework leads to a similar training paradigm, it offers a fresh perspective from diffusion models. This introduces new design spaces, including loss weight considerations. Additionally, our sampling approach is more flexible, allowing for both implicit timestep and explicit timestep sampling.

\paragraph{Discrete-state diffusion models for discriminative tasks.}\citet{liu2023pyramid} apply these models to 3D scenes, \citet{wang2023segrefiner} investigates segmentation refinement, and \citet{inoue2023layoutdm} employs them for layout generation. We provide in-depth analysis between Masked Generative Models and Diffusion Models under various noise schedules within the broader framework of Discrete Interpolants, and frame segmentation as unmasking from \mmm tokens. Furthermore, we jointly train on image-segmentation mask pairs to model the joint distribution by Discrete Interpolants, enabling both image-conditioned mask generation and mask-conditioned image generation.

\section{Method}
\label{sec:flow}

\begin{table}
    \centering
        \resizebox{0.98\linewidth}{!}{
    \begin{tabular}{l|c|c}
    \toprule
    \textbf{Name}& \textbf{Equation} $\kappa_t$ & \textbf{Derivative} $\dot{\kappa}_t$  \\
    \hline 
    Root~\cite{weber2024maskbit} & $\sqrt{t}$  & $\frac{1}{2\sqrt{t}}$  \\
    \hline 
Linear~\cite{campbell2024generative}   & $t$  & $1$  \\
      \hline 
      Cosine~\cite{chang2022maskgit}   &  $1-cos(\frac{\pi t}{2})$ & $\frac{\pi}{2} \sin\left(\frac{\pi t}{2}\right)$\\
      \hline 
      Arccos~\cite{weber2024maskbit} & $1-\frac{2 arccos(t)}{\pi}$ & $\frac{2}{\pi \sqrt{1-t^2}}$\\
      \hline 
      {Quadratic} & $t^{2}$ & $2t$  \\
      \hline 
      Cubic~\cite{gat2024discrete} & \makecell{ $-2t^3 + 3t^2 + b(t^3-t^2)$ \\ $a(t^3 -2t^2+t)$} & \makecell{$(-6 + 3a + 3b)t^2$\\$ + (6 - 4a - 2b)t + a$ } \\
      \hline 
       C Coupling~\cite{gat2024discrete} & $\bar{\kappa}_t\cdot \delta_{i}(x_t) + \bar{\kappa}_0\cdot (1- \delta_{i}(x_t))  $ & $--$  \\
         \bottomrule
    \end{tabular}
    }
    \caption{\textbf{Different Masking Schedules $\mathbf{\kappa_t}$.} These can be applied in Discrete Interpolants: $ p_{t|0,1}(x|x_0,x_1) = (1-\kappa_t) \delta_{x_0}(x) +  \kappa_t \delta_{x_1}(x) $. ``C Coupling" refers to Conditional coupling as described in~\citet{gat2024discrete}, it can be seen as a token-dependent scheduler, for more details please refer to Appendix.
    }
    \label{tab:scheduler}
\end{table}

\subsection{Discrete  Interpolants}

Our method draws inspiration from discrete-state Diffusion/Flow models~\cite{gat2024discrete,ou2024absorbingdiscretediffusionsecretly_radd,sahoo2024simple,zheng2024masked} and continuous Stochastic Interpolants~\cite{albergo2022building,albergo2023stochastic,ma2024sit}. We aim to extend this interpolant method into discrete-state models with the creation of a flexible and scalable framework for discrete-state modeling.
 Given the $L$-length real data  $x_1 \in \mathbb{R}^{L}$, the entire possible set is defined as $\mathcal{D} = [K]^{L}$, where $[K]={1,2,...,K}$, and $K$ is the vocabulary size (including an extra \mmm token). $x_0 \in \mathbb{R}^{L}$ represents noise, typically tokens filled with mask token $\mm$. %
 
Our goal is to learn a transition process based on a vector field $u(x_t,t)$ from $x_0$ (fully masked) to real data $x_1$ by progressively unmasking tokens at each timestep $t$. The key process is a Flow Matching-style probability path interpolated according to the masking schedule $\kappa_t$~\footnote{Various terms are used to describe interpolants across different papers, such as ``noise schedule" in diffusion models, ``interpolants" in~\cite{albergo2023stochastic}, ``protocol" in~\cite{shih2022training}, and ``scheduler" in~\cite{gat2024discrete}. For simplicity, we consistently refer to this concept as ``scheduler" throughout our paper.}:

\begin{equation}
\label{eq:simple_interpolants}
p_{t|0,1}(x|x_0,x_1) = (1-\kappa_t) \delta_{x_0}(x) + \kappa_t \delta_{x_1}(x),
\end{equation}
where $\delta_{x_0}(x)$ can be instantialized as $\delta_{\mm}(x)$, and  $\delta_{x}(\cdot)$ is a Dirac delta function indicating whether the element is $x$. $\kappa_t \geq 0$ and $\kappa_0 = 0$ and $\kappa_1 = 1$. There are several choices for the masking schedule $\kappa_t$ , e.g., linear, cosine, etc, as shown in~\cref{tab:scheduler}.

\subsection{Training}

In the case of continuous-state, given a probability path $p(x_t)$ and a vector field $u(x_t,t)$, the key to ensuring that traversal along the vector field $u(x_t,t)$ can yield the probability transition between $p(x_0)$ and $p(x_1)$ is the Continuity Equation~\cite{song2021scorebased_sde}.
Similarly, in discrete-state modeling, there's a counterpart theory called the Kolmogorov Equation \cite{campbell2024generative}. It indicates that by following the design of $u_t(x_t) = \frac{\dot{\kappa}_t}{1-\kappa_t}[{p_{1|t}(x_1|x_t,t;\theta)-\delta_{x_t}(x)}]$, we can traverse along the vector field $u(x_t,t)$ to yield the probability transition between $p(x_0)$ and $p(x_1)$. To learn such a vector field $u_t(x_t) = \frac{\dot{\kappa}_t}{1-\kappa_t}[{p_{1|t}(x_1|x_t,t;\theta)-\delta_{x_t}(x)}]$, we only need to learn $p^{\theta}_{1|t}(x_1|x_t,t)$, which incidentally acts as an unmasking function, and  can be optimized by a cross-entropy loss:

\vspace{-0.5cm}
\begin{multline}
\label{eq:elbo_loss}
\mathcal{L}(\theta) = \E_{\pdata(\x_1)p(\x_0) \mathcal{U}(t; 0, 1) p_{t|0,1}(\x_t | \x_0,\x_1)}\\
\log \denoise(\x_1 | \x_t,t;\theta),
\end{multline}
\vspace{-0.5cm}

where $x_t$ is obtained by Discrete Intepolants in~\cref{eq:simple_interpolants}.

\begin{figure*}
    \centering
    \includegraphics[width=0.9\linewidth]{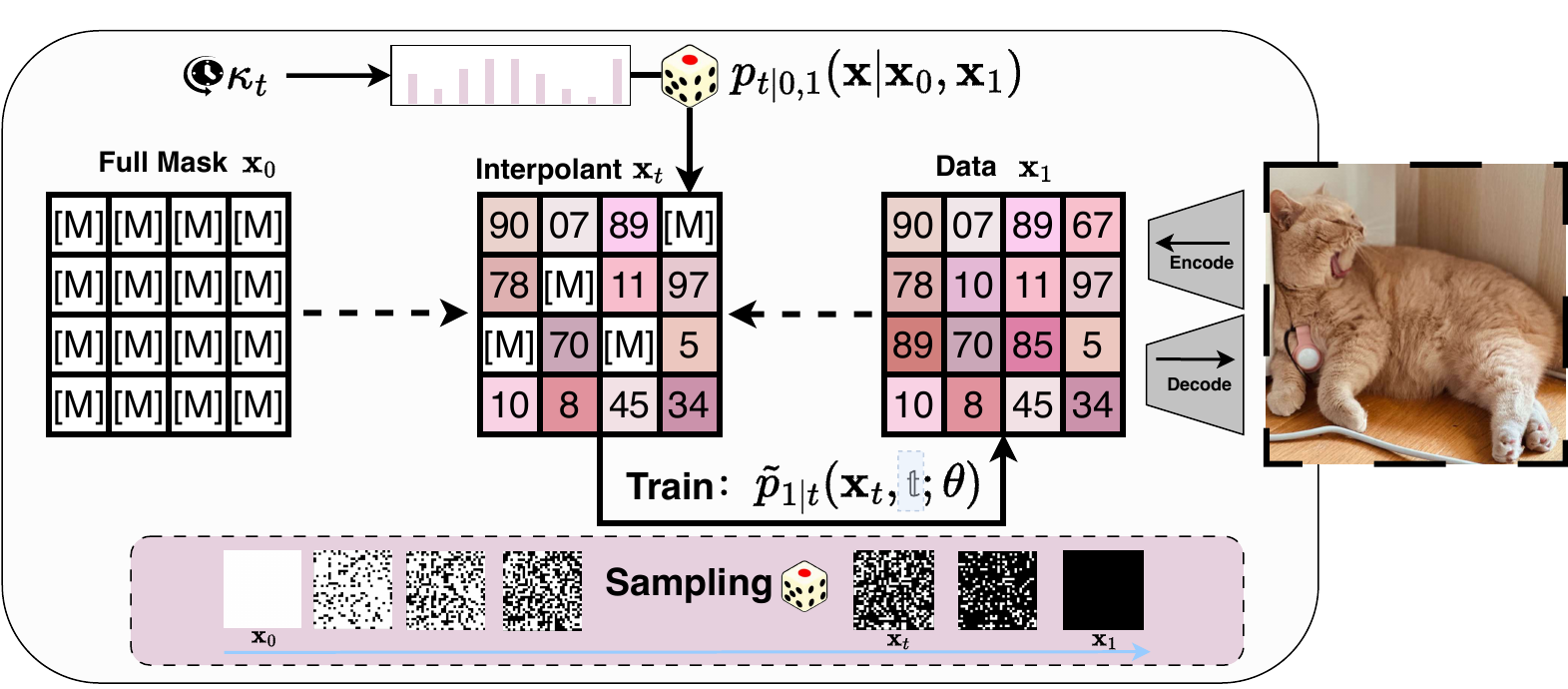}
    \caption{\textbf{Discrete Interpolants for training and sampling:} During training, we first obtain discrete interpolants $x_t$ from $x_0$ and $x_1$ following a specific scheduler $\kappa_t$. We then train a model with the cross-entropy loss to predict the original data with $\tilde{p}_{1|t}(\textbf{x}_t,\mathbbm{t};\theta)$, where $\mathbbm{t}$ indicates that our timestep $t$ is optional, leading to both Explicit Timestep and Implicit Timestep Models. For sampling, we begin with a fully masked $x_0$ and progressively unmask to reach the final fully unmasked $x_1$. Lastly, we decode the indices back to pixel space.
    }
    \label{fig:method}
\end{figure*}

\paragraph{Masking and Weighting on Cross-Entropy.}
To improve the performance of fidelity, and stabilize the training,  we further introduce two significant modifications upon to cross-entropy formulation: a masking operation in the cross-entropy loss and a weighting mechanism:

\vspace{-6mm}
\begin{multline}
\label{eq:main}
\mathcal{L}(\theta) = \E_{\pdata(\x_1)p(\x_0) \mathcal{U}(t; 0, 1) p_{t|0,1}(\x_t | \x_0,\x_1)} \\
\left[ \underbrace{\textcolor{red}{w(t)}}_{\text{\textcolor{red}{Weighting}}} \underbrace{\textcolor{blue}{\delta_{\mm}(x_t) (x_1)^\top}}_{\text{\textcolor{blue}{Masking}}}  \log \denoise(\x_1 | \x_t,t;\theta) \right],
\end{multline}

where $w(t)$ is the weighting function across various timesteps, as indicated by~\citet{kingma2021variational}, the detailed form of $w(t)$ indicates the weighted timestep integral, normally  $w(t)=\frac{\dot{\kappa}_t}{1-\kappa_t}$ for the ELBO (Evidence Lower Bound) loss~\cite{kingma2024understanding}, provided the start and end points of the time schedule $\kappa_t$ remain unchanged. However, since ELBO optimization primarily targets improved log-likelihood rather than visual quality, an appropriate $w(t)$ is necessary to enhance visual output. 
We empirically validate that this holds in the discrete-state setting as well.
Unlike BERT~\cite{kenton2019bert} and~\citet{gat2024discrete}, which do not incorporate masking, we found that applying masking within the cross-entropy loss is crucial for mitigating overfitting in vision domain.

\paragraph{From Explicit Timestep Model to Implicit Timestep Model:~\boldsymbol{$p_{1|t}(x_1|x_t,t;\theta)\rightarrow p(x_1|x_t;\theta)$}.} 
Another interesting property of our method is that $p_{1|t}(x_1|x_t,t;\theta)$ can evolve from an Explicit Timestep Model into an Implicit Timestep Model when considering masked modeling, by removing the dependence of timestep $t$, leading to an unmasking model $p(x_1|x_t;\theta)$. In detail, recent works~\cite{ou2024absorbingdiscretediffusionsecretly_radd,sahoo2024simple,zheng2024masked,shi2024simplified} have demonstrated that we can remove the dependence on timestep in model design. They show that timestep-dependence can be extracted as a form of weight coefficient outside of the cross-entropy loss. Assume that our scheduler $\kappa_t$ is reversible, even with strict monotonicity, a more intuitive interpretation is that, given a set of masked data by a specific scheduler $\kappa_t$ in $t$, this masked data inherently contains information about which timestep $t$ the data $x_t$ is from (i.e., how corrupted the data $x_t$ is). Thus, we can safely remove the dependence on the timestep in the model\footnote{In continuous-state diffusion models, the corrupted data also implicitly includes timestep information, but it's not as straightforward as in discrete-state models to determine the level of corruption in the current data $x_t$.}. This implicit timestep property offers several advantages:
1). It establishes a close connection with Masked Generative Model,e.g.,  MaskGiT~\cite{chang2022maskgit}, though with a key difference: our discrete model employs independent sampling per token based on pure probability, whereas MaskGit's sampling is heuristically based on the confidence score per token. In experiment, we show that our method can also deploy MaskGit-style sampling but training by our diffusion-based paradigm.
2). The sampling steps can be much simpler and are upper-bounded by the token length $L$.
3). The extra explicit timestep $t$ can sometimes be a restriction in special scenarios, e.g., specific-order sampling or image editing. In these cases, we would prefer to sample auto-regressively or row-by-row. Thus, it's challenging to define the explicit timestep; instead, we could use the implicit timestep from $x_t$ itself.

\paragraph{From Single-Token to Multi-Token.}
In  previous part, we mainly consider single token and introduce the interpolants, and training process of them. In real scenario, data $x$ can be a $L$-length token sequence $x^{i},i \in [L]$. We assume the interpolants operation  can be factorized as token-wise. The network predicts the probabilities of all tokens at a time $t$. The loss in~\cref{eq:main} under the multi-token case can be written as:

\vspace{-5mm}
\begin{multline}
\label{eq:elbo_loss_multi_token}
\mathcal{L}(\theta) = \E_{\pdata(\x_1)p(\x_0) \mathcal{U}(t; 0, 1) p_{t|0,1}(\x_t | \x_0,\x_1)}  \\
\left[ w(t)\sum_{l:x_t^{l}=\mm} (x_1^{l})^\top  \log \denoise^l(\x_1 | \x_t,t;\theta) \right].
\end{multline}

Given that sampling is independent for each token, and interactions between tokens occur only within the network parameterized by $\theta$. This shares the same expression as previous works~\cite{shi2024simplified,gong2024scaling} with masking and weighting. We'll express the sampling using the formulation of the single-token case, omitting multi-token considerations (the superscript $l$), to enhance clarity and understanding.

\subsection{Sampling}

During sampling process, the sample $x_t$ progressively changes between states in $\mathcal{D}= [K]^{L}$ during sampling, aiming to unmask every token of $x$ with step size $\Dt$.  Our framework offers flexibility in sampling types. We primarily consider three types:
\begin{itemize}
    \item Explicit Timestep Model: This model design is timestep-dependent as conventional diffusion models.
    \item Implicit Timestep Model: This model design is timestep-independent  by removing timestep $t$ in the backbone.
    \item Masked Generative Model's style sampling following a greedy heuristic manner: Based on the Implicit Timestep Model, we explore the conventional greedy heuristic style from Masked Generative Models, such as MaskGit~\cite{chang2022maskgit}. Our unmasking network functions similarly to Masked Generative Models by recovering masked tokens. We can use our pretrained unmasking network for sampling, following MaskGiT's procedures with $u_t(x_t) = p(x_1|x_t,t;\theta)$ as the token unmasking function.
\end{itemize}

We detail the sampling process of  Explicit and Implicit Timestep Models in~\cref{alg:sampling}. For the third type, we directly follow MaskGit's pipeline, which we omit from our sampling algorithm, while still using our pretrained timestep-independent unmasking network $u_t(x_t) = p(x_1|x_t;\theta)$.

 Notably, unlike the greedy heuristic rule by purity~\cite{tang2022improved_vqdiffusion} or confidence~\cite{chang2022maskgit}, which can guarantee all tokens will be unmasked till the next step, in our model sampled by diffusion models, this is not always true, especially when the sampling step is too small, see~\cref{fig:argmax_save_scheduler}. So we add one extra \texttt{argmax} operation on the logits space of the output in the last step of the sampling to mitigate this issue to ensure all tokens will be unmasked. This operation can effectively churn the sampling process to efficiently achieve the data point in a high-fidelity manner.

\begin{algorithm}
    \caption{\textbf{Sampling process of Implicit or Explicit Timestep Model with fixed step size.} 
    }
    \begin{algorithmic}[!ht]
    \REQUIRE Network $p(x_1|x_t;\theta)$, masking schedule $\kappa_t$, mask token $m$, time range  $t \in [0,1]$, step size $\Delta t$    
    \STATE  $t \gets 0$, 
       $\x_0 \gets m$, 
        \WHILE{$t <=1 - \Delta t$} 
             \STATE $u_t(x_t) = \frac{\dot{\kappa}_t}{1-\kappa_t}({p (x_1|x_t,t;\theta)-\delta_{x_t}(x)})$
            \STATE \scalebox{0.9}{$p(x_1|x_{t+\Delta t},t+\Delta t;\theta) \gets \text{Cat}\left[\delta_{x_t}(t+ \Dt) + u_t(x_t)  \Dt\right]$}
            \STATE $\x^i_{t+ \Dt} \sim p^i(x_1|x_{t+\Delta t},t+\Delta t;\theta)$  for all $x_t^i=m$, $x_{t+\Delta t}^i  \gets x_t^i$ for all $x_t^i \neq m$.
            \STATE $t  \gets t + \Delta t$
        \ENDWHILE
        \STATE $x^i_1 = \texttt{argmax}  \hspace{3mm} u_t^i(x_t) $ for all $x_t^i=m$
    \end{algorithmic}
    \label{alg:sampling}
\end{algorithm}

\subsection{Segmentation is Unmasking}

A natural extension of Discrete Interpolants is to consider multimodal joint learning the joint distribution, inspired by~\cite{assran2023self} as well as recent advances in combined discriminative and generative learning~\cite{chen2024denoisingjointembeddingpredictivearchitecture,li2023mage}.
Given real image $x_1 \in \mathbb{R}^{L_x}$ and the second modality $y_1 \in \mathbb{R}^{L_y}$, where $L_x$, $L_y$ is the token number of the respective modalities(note here, by default the token dimension of the image should be 2D dimensions, we assume that we have squeezed them into single dimension for better illustration). 
We simply feed them into the network and obtain their specific logits, therefore, we formulate the unified loss as:

\vspace{-4mm}
\begin{multline}
 \mathcal{L} (\theta) = \E_{\pdata(\x_1,y_1) \mathcal{U}(t; 0, 1) \noisemarg(\x_t | \x_1) \noisemarg(\y_t | \y_1)} \\
\left[ w(t) \delta_{\mm}(z_t) (z_1)^\top  \log \denoise(z_1 | z_t,t;\theta) \right],
\end{multline}

where $z_1=\x_1\oplus\y_1,z_t=\x_t\oplus\y_t$, $\oplus$ is the concatenation operation after flattening, and $p_{1|t}(z_1 | z_t;\theta)$ is the unmasking process in discrete models by simply replacing real token by \mm according to the masking schedule $\kappa_t$. Noticeably, we share the mask schedule between two different modalities, we find it empirically works well. For the parameterization of $\theta$,  the core idea is to standardize each task’s inputs and outputs as sequences of discrete vocabulary tokens. To better leverage inductive bias, we retain the patchification approach used in U-ViT~\cite{uvit}. By default, our method operates in latent space to reduce the token count effectively.

\subsection{Classifier-free Gudiance}

We can conduct versatile sampling processes in both single-modality and double-modality scenarios. For the sake of generality, we'll focus on the conditional sampling of $p(x|y;\theta)$ in double-modalities:

\vspace{-8mm}
\begin{multline}
\label{eq:cfg}
    p(x_1|x_t,y;\omega,\theta) = \\
    p(x_1|x_t; \cfg,\theta) + \omega \left[p(x_1|x_t,y;\theta)  - p(x_1|x_t,\cfg;\theta) \right],
\end{multline}
where $\omega$ represents the guidance strength, and  token \cfg serves as the null embedding to indicate unconditional signal. By swapping the positions of x and y, we can achieve a similar classifier-free guidance for $p(y|x;\theta)$. This guidance  can serve as a plug-in replacement for our previous sampling function described in ~\cref{alg:sampling}.

\subsection{Extra Discussions}

\paragraph{Connection between Cold Diffusion~\cite{bansal2024colddiffusion}.} Cold Diffusion demonstrates a generalized diffusion model, that can be composed of Degradation Operator $\mathcal{D}$ and Restore Operator $\mathcal{R}$, the degradation can be blurring, pixelated, or snowification, with restriction $\mathcal{D}(x_1,1)=x_1$, their training is achieved by a minimal optimization problem:
\begin{equation}
    \text{min}_{\phi} \mathbb{E}_{x}||\mathcal{R}(\mathcal{D}(x,t),t;\phi) -x||,
\end{equation}
our masking operation in~\cref{eq:simple_interpolants} can be naturally a case of the degradation $\mathcal{D}(x_t,t)\sim p_{t|0,1}(x|x_0,x_1)$, and our unmasking network is a case of the Restore Operator $\mathcal{R}(x_t,t)$.

\paragraph{Conditional Coupling is secretly a token-dependent scheduler.} Conditional Coupling~\cite{gat2024discrete} is a method for coupling $x_0$ and $x_1$ when constructing the discrete interpolants: $x_0=\mathbb{I} m,\mathbb{I} m,..,(1-\mathbb{I})\x_1, (1-\mathbb{I})\x_1 $, it can be seen as a special case of the masking schedule:

\vspace{-3mm}
\begin{equation}
\label{eq:condition_coupling}
p_{t|0,1}(x|\bar{x}_0,\bar{x}_1) = (1-\kappa_t) \delta_{\bar{x}_0}(x) + \kappa_t \delta_{\bar{x}_1}(x),
\end{equation}
where $(\bar{x}_0,\bar{x}_1) = (\left[ \mathbb{I} \otimes x_1 + (1-\mathbb{I})\otimes (\mm,...,\mm) \right],x_1 )$. It can be understood as a special scheduler built upon the default by making the scheduler dependent on the token's location, $\kappa_t\rightarrow \kappa^i_t$. If $i \notin \mathbb{I}$, it conducts a standard interpolant based on the scheduler; if $i \in \mathbb{I}$, the scheduler is a null scheduler, meaning the token remains unchanged. $\mathbb{I}$ is a mask controlled by a ratio of data length $L$, determining the balance between standard and null schedulers.

\section{Experiment}

\subsection{Experimental Detail}

\paragraph{Datasets and Metrics.} For image generation, we primarily focus on ImageNet256 and MS COCO datasets. For joint training between image-segmentation mask pairs, we use  the Cityscapes dataset. Our video generation experiments mainly utilize the FaceForensics dataset.
We use Fréchet Inception Distance (FID)  as the evaluation metric for image generation tasks. For video generation, we use Fréchet Video Distance (FVD) to assess performance.

\paragraph{Training Details.} We consistently use the discrete tokenizer SD-VQ-F8 from Stable Diffusion~\cite{rombach2022high_latentdiffusion_ldm} across all datasets, due to its extensive pretraining on large datasets like Open-Images~\cite{kuznetsova2020openimages}. To avoid singularity issues in the derivative, we sample timesteps $t \in [\epsilon, 1-\epsilon]$ during training, where $\epsilon = 10^{-3}$. For fair comparison, we employ the same scheduler for both training and sampling, using a fixed step size. For classifier-free guidance, we randomly drop out the conditional signal with a probability of 0.1 during training, following established conventions~\cite{ho2021classifier}.  Unlike ~\cite{gat2024discrete}, which uses adaptive step size for sampling, we consistently employ a fixed step size for fair comparison. Unless otherwise stated, we default to using 1{,}000 timesteps. For COCO, ImageNet, and Faceforensics, we use linear schedules by default and set $w(t)=1$, and applying the masking cross-entropy. For more detailed information about the training recipe, optimizer, iteration steps, GPU usage, learning rate, and other parameters, please refer to the Appendix.

\begin{table*}
    \centering
    \scalebox{0.76}{
    \begin{tabular}{lcccc}
    \toprule
        \textbf{Model} & \textbf{FID} & \textbf{Type} & \textbf{Training datasets} & \textbf{\#Params} \\
    \midrule
    \multicolumn{4}{l}{Generative model trained on external large dataset (zero-shot)} \\
    \arrayrulecolor{black!30}\midrule
        \quad \textcolor{gray}{LAFITE}~\cite{zhou2021lafite} & \textcolor{gray}{26.94} & \textcolor{gray}{GAN} & \textcolor{gray}{CC3M (3M)} & \textcolor{gray}{75M + 151M (TE)}  \\
        \quad \textcolor{gray}{Parti}~\cite{yu2022scaling_parti} & \textcolor{gray}{7.23} & \textcolor{gray}{Autoregressive} & \textcolor{gray}{LAION (400M) + FIT (400M) + JFT (4B)} & \textcolor{gray}{20B + 630M (AE)} \\
        \quad \textcolor{gray}{Re-Imagen}~\cite{chen2022re_imagen} & \textcolor{gray}{6.88} & \textcolor{gray}{Continous Diffusion} & \textcolor{gray}{KNN-ImageText (50M)} & \textcolor{gray}{2.5B + 750M (SR)} \\
    \arrayrulecolor{black}\midrule
    \multicolumn{4}{l}{Generative model trained on external large dataset with access to MS-COCO} \\
    \arrayrulecolor{black!30}\midrule
            \quad \textcolor{gray}{Re-Imagen}$^\ddagger$~\cite{chen2022re_imagen} & \textcolor{gray}{5.25} & \textcolor{gray}{Diffusion} & \textcolor{gray}{KNN-ImageText (50M)} & \textcolor{gray}{2.5B + 750M (SR)} \\
             \quad Make-A-Scene~\cite{gafni2022make_a_scene} & 7.55 & Autoregressive & Union datasets (with MS-COCO) (35M) & 4B \\
        \arrayrulecolor{black!30}\midrule
         \quad VQ-Diffusion$^\dagger$~\cite{vqdiffusion} & 13.86 & Discrete diffusion & Conceptual Caption Subset (7M) &  370M \\
    \arrayrulecolor{black}\midrule
    \multicolumn{4}{l}{Generative model trained on MS-COCO} \\
    \arrayrulecolor{black!30}\midrule
       
        \quad \textcolor{gray}{U-Net} & \textcolor{gray}{7.32} & \textcolor{gray}{Continuous diffusion} & \textcolor{gray}{MS-COCO (83K)} & \textcolor{gray}{53M + 123M (TE) + 84M (AE)} \\
        \quad \textcolor{gray}{U-ViT}~\cite{uvit} & \textcolor{gray}{5.48} & \textcolor{gray}{Continuous diffusion} & \textcolor{gray}{MS-COCO (83K)} & \textcolor{gray}{58M + 123M (TE) + 84M (AE)} \\
        \arrayrulecolor{black!30}\midrule
         \quad VQ-Diffusion~\cite{vqdiffusion} & 19.75 & Discrete Diffusion & MS-COCO (83K) & 370M \\
         \hline
         \quad Implicit Timestep Model (\textit{Our},w/ 20-steps) & \text{8.11}  & Discrete Diffusion~\&~MGM & MS-COCO (83K) & 77M  + 123M (TE) + 84M (AE)  \\
        \quad Implicit Timestep Model (\textit{Our}) & \textbf{5.65}  & Discrete Diffusion~\&~MGM & MS-COCO (83K) & 77M  + 123M (TE) + 84M (AE)  \\
        \quad Explicit Timestep Model (\textit{Our}) & \textbf{6.03}  & Discrete Diffusion~\&~MGM & MS-COCO (83K) & 77M  + 123M (TE) + 84M (AE)  \\
    \arrayrulecolor{black}\bottomrule
    \end{tabular}
    }
        {\\\fontsize{5}{\baselineskip}\selectfont SR represents a super-resolution module, AE an image autoencoder, and TE a text encoder. Methods marked with $^\dagger$ are finetuned on MS-COCO. Those marked with $^\ddagger$ use MS-COCO as a knowledge base for retrieval.}
    \vspace{-2mm}
     \caption{
 \textbf{FID results of different models on MS-COCO ($256 \times 256$).} MGM denotes Masked Generative Models.  Baseline results are sourced from U-ViT~\cite{uvit}.
}
 \label{tab:coco}
\end{table*}

\begin{figure}
    \centering
    \includegraphics[width=0.99\linewidth]{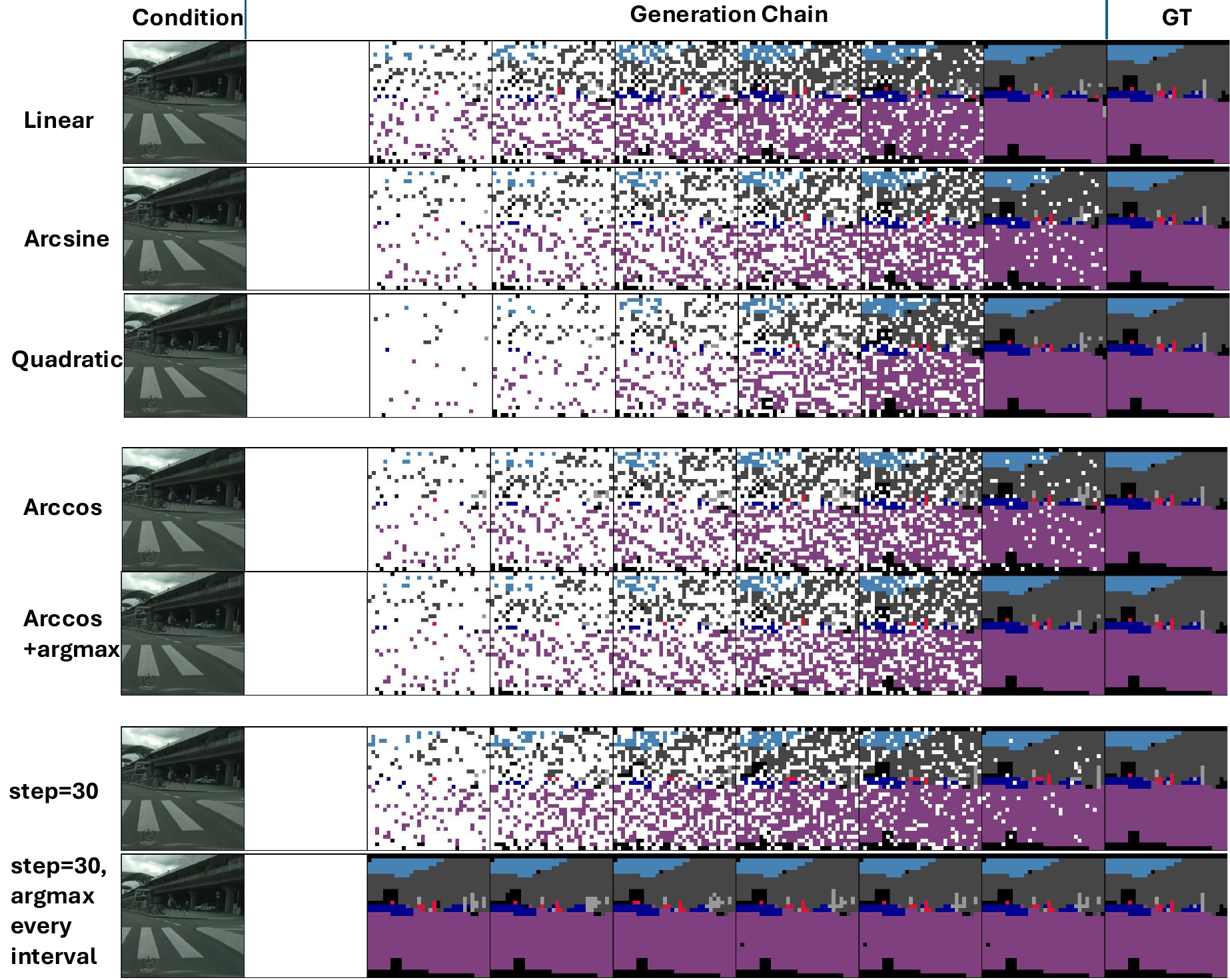}
    \caption{
    \textbf{Churning sampling by \texttt{argmax} can 1), alleviate the misalignment between schedulers. 2), boost sampling performance in low-NFE.} First, we visualize the progressive chain of changes when sampling with a scheduler $\kappa_t$ that differs from the linear scheduler used during training. Our sampling process uses 50 steps and a CFG scale of 3. Second, we demonstrate that applying the \texttt{argmax} operation to logits can significantly reduce the occurrence of remaining \mmm tokens after sampling.
    }
    \label{fig:argmax_save_scheduler}
\end{figure}

\begin{table}
    \centering
    \resizebox{0.98\linewidth}{!}{
    \begin{tabular}{c|l|ccc}
    \toprule
    \textbf{Type} & \textbf{Model} & \textbf{\#Para.} & \textbf{FID}$\downarrow$ & \textbf{IS}$\uparrow$ \\
    \midrule
    \multirow{9}{*}{\makecell{AR \& MGM \\ Models}} %
     & VQGAN~\citep{esser2021taming_vqgan}    & 1.4B   & 15.78 & 74.3      \\
     & VQGAN-re~\citep{esser2021taming_vqgan}  & 1.4B  & 5.20  & 280.3    \\
     & ViT-VQGAN~\citep{yu2021vector_vitvq} & 1.7B & 4.17  & 175.1        \\
     & ViT-VQGAN-re~\citep{yu2021vector_vitvq}& 1.7B  & 3.04  & 227.4     \\
     & RQTran.~\citep{lee2022autoregressive_rqtran}       & 3.8B  & 7.55  & 134.0    \\
     & RQTran.-re~\citep{lee2022autoregressive_rqtran}    & 3.8B & 3.80  & 323.7     \\
      & LlamaGen-XL~\citep{sun2024autoregressive_llamagen} & 775M & 3.39 & 227.1\\
    & MaskGIT~\cite{chang2022maskgit} & 227M & 6.18 & 182 \\
     & Open-MAGVIT2-L~\cite{luo2024openmagvit2} & 804M & 2.51 & 271.7  \\
         \midrule
      \multirow{4}{*}{\makecell{Continuous \\Diffusion} } & ADM~\citep{dhariwal2021diffusion_beat}  & \textcolor{black}{554M}       & \textcolor{black}{10.94} & \textcolor{black}{101.0}        \\
     & CDM~\citep{ho2022cascaded}   & \textcolor{black}{$-$}       & 4.88  & \textcolor{black}{158.7}         \\
     & \textcolor{black}{LDM-4}~\citep{rombach2022high_latentdiffusion_ldm} & \textcolor{black}{400M}     & \textcolor{black}{3.60}  & \textcolor{black}{247.7}       \\
     & \textcolor{black}{DiT-XL/2}~\citep{dit_peebles2022scalable}  & \textcolor{black}{675M}  & \textcolor{black}{2.27}  & \textcolor{black}{278.2}      \\
    \midrule
    Discrete 
    &  VQ-Diffusion~\cite{vqdiffusion}& & 5.32 & $-$  \\
     \hline 
      \hline 
     DD \& MGM &  Explicit Timestep Model(\textit{Our}) & 546M & 5.84 & {186.1} \\
     DD \& MGM &  Implicit Timestep Model(\textit{Our}) & 546M &  \textbf{{5.30}}  & {183.0}   \\
    \bottomrule
    \end{tabular}
    }
    \caption{\textbf{Class-conditional generation on  ImageNet $256 \times 256$.} MGM denotes Masked Generative Models. AR denotes Auto-Regressive Models. DD denotes Discrete Diffusion Models.
    }
     \label{tab:in256}
\end{table}

\begin{figure*}
    \centering
    \begin{subfigure}{0.33\textwidth}
        \includegraphics[width=\linewidth]{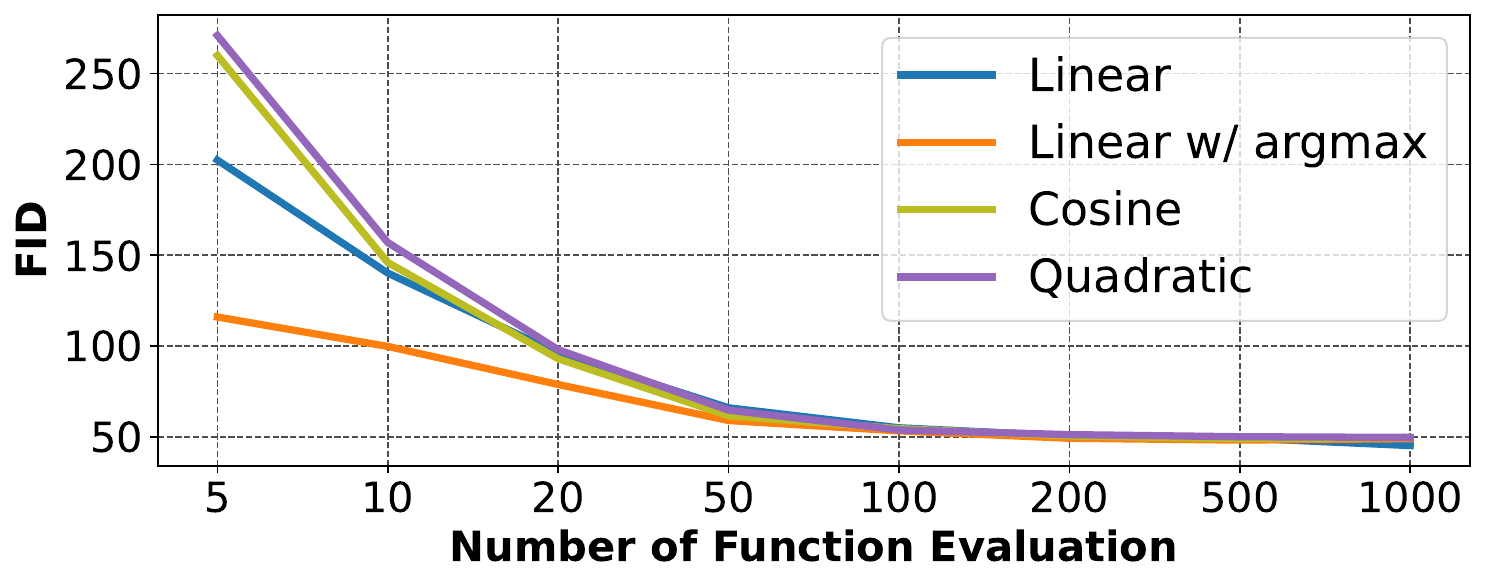}
        \caption{NFE in ETM.}
        \label{fig:etm_nfe_fid}
    \end{subfigure}
    \hfill
    \begin{subfigure}{0.33\textwidth}
        \includegraphics[width=\linewidth]{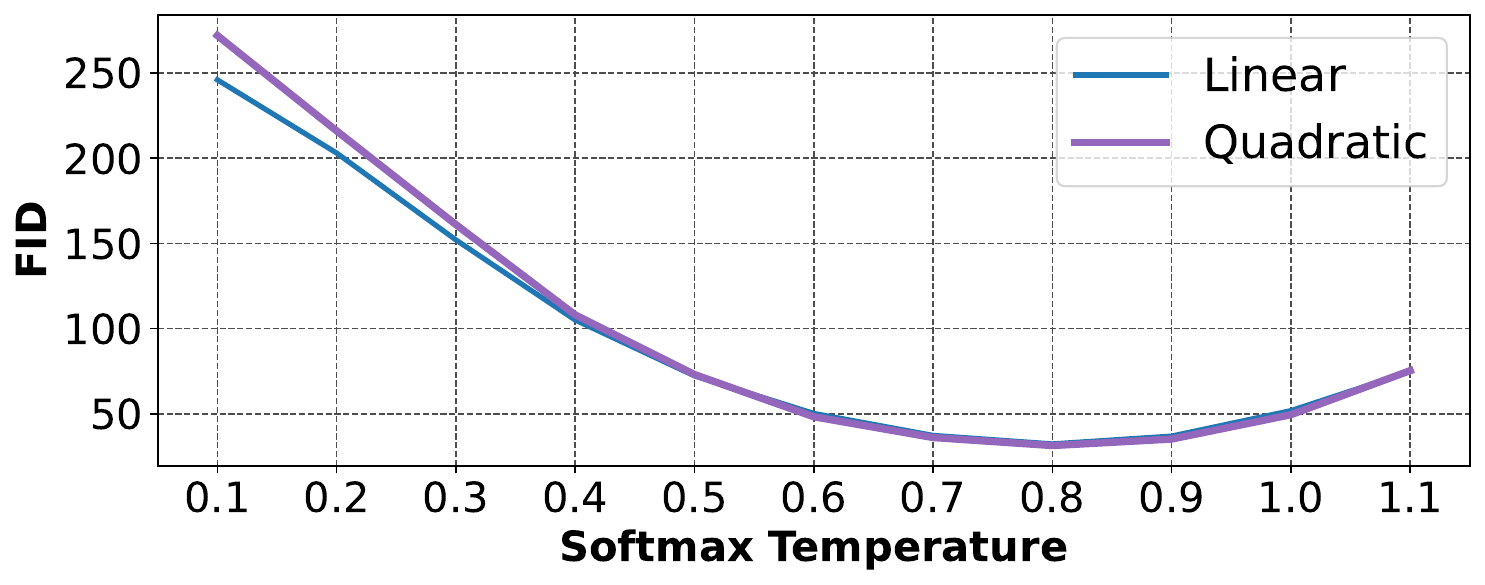}
        \caption{Temperature in ETM.}
        \label{fig:etm_temp_fid}
    \end{subfigure}
    \hfill
    \begin{subfigure}{0.33\textwidth}
        \includegraphics[width=\linewidth]{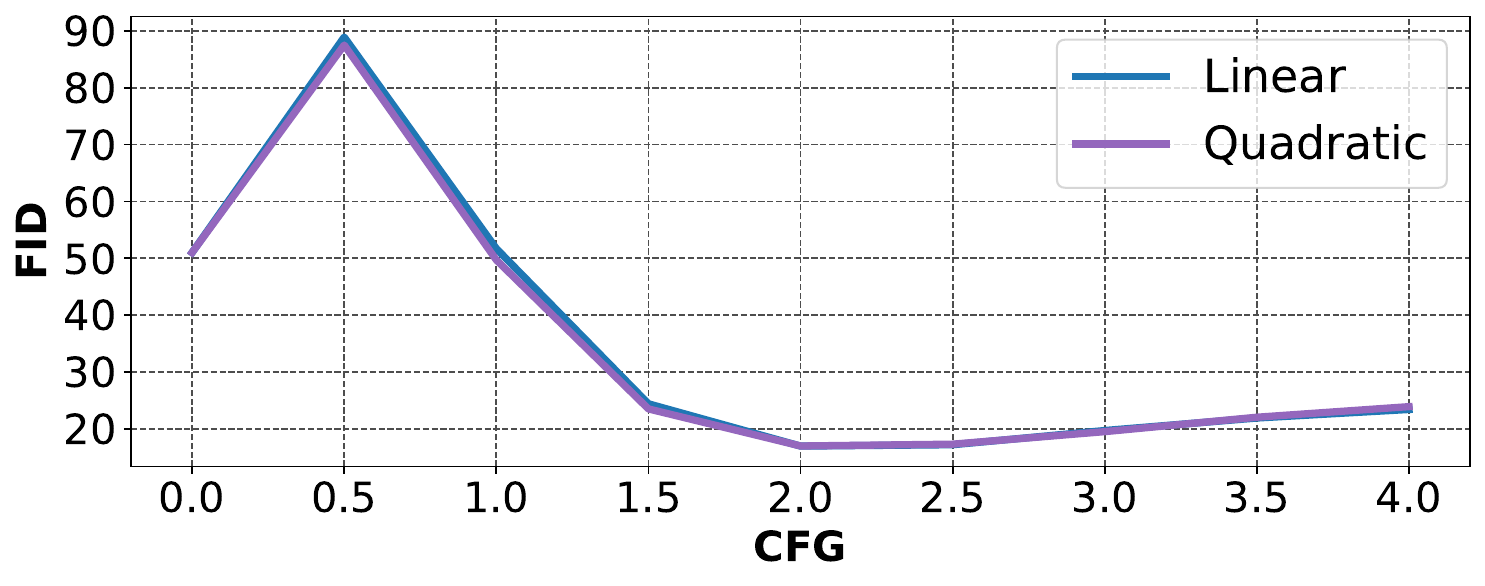}
        \caption{CFG in ETM.}
        \label{fig:etm_cfg_fid}
    \end{subfigure}

    \vspace{0.2cm} %

    \begin{subfigure}{0.33\textwidth}
        \includegraphics[width=\linewidth]{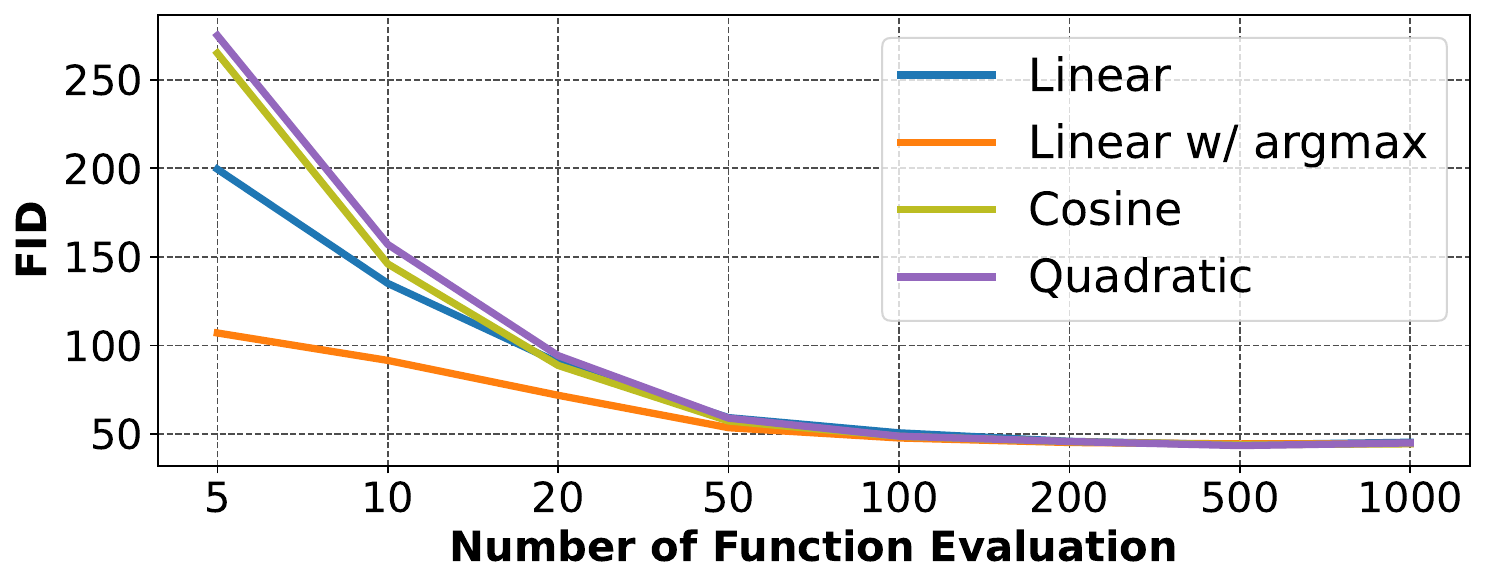}
        \caption{NFE in ITM.}
        \label{fig:itm_nfe_fid}
    \end{subfigure}
    \hfill
    \begin{subfigure}{0.33\textwidth}
        \includegraphics[width=\linewidth]{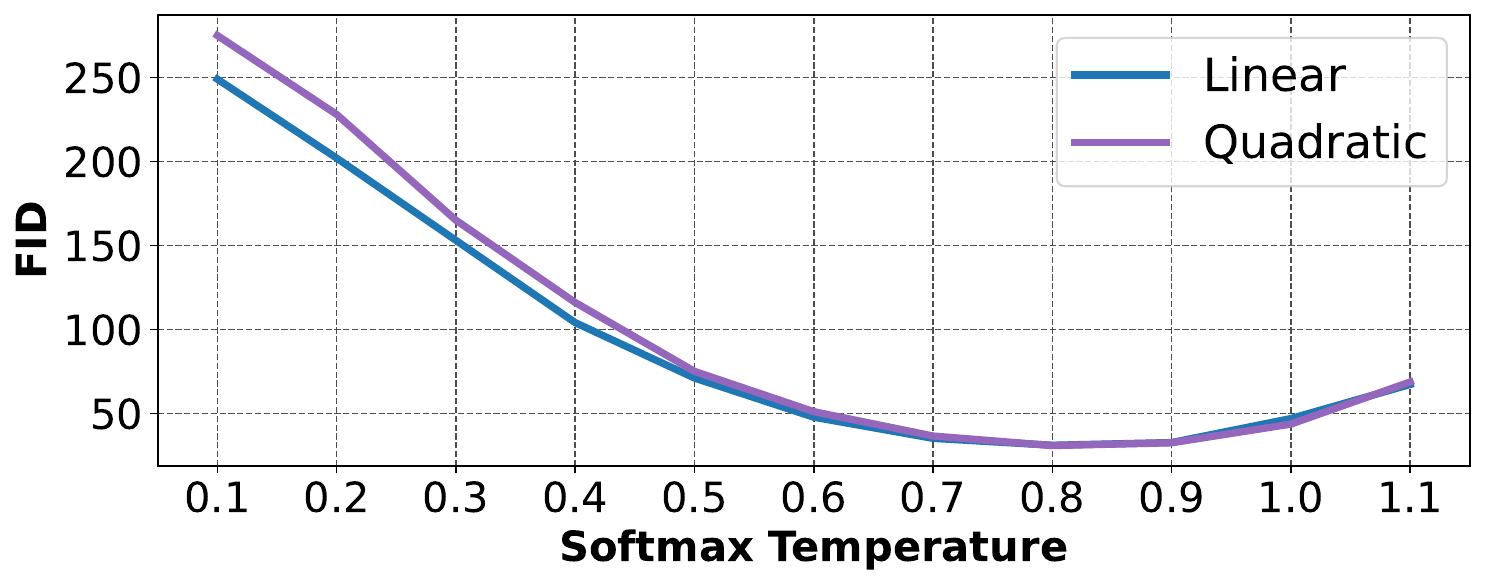}
        \caption{Temperature in ITM.}
         \label{fig:itm_temp_fid}
    \end{subfigure}
    \hfill
    \begin{subfigure}{0.33\textwidth}
        \includegraphics[width=\linewidth]{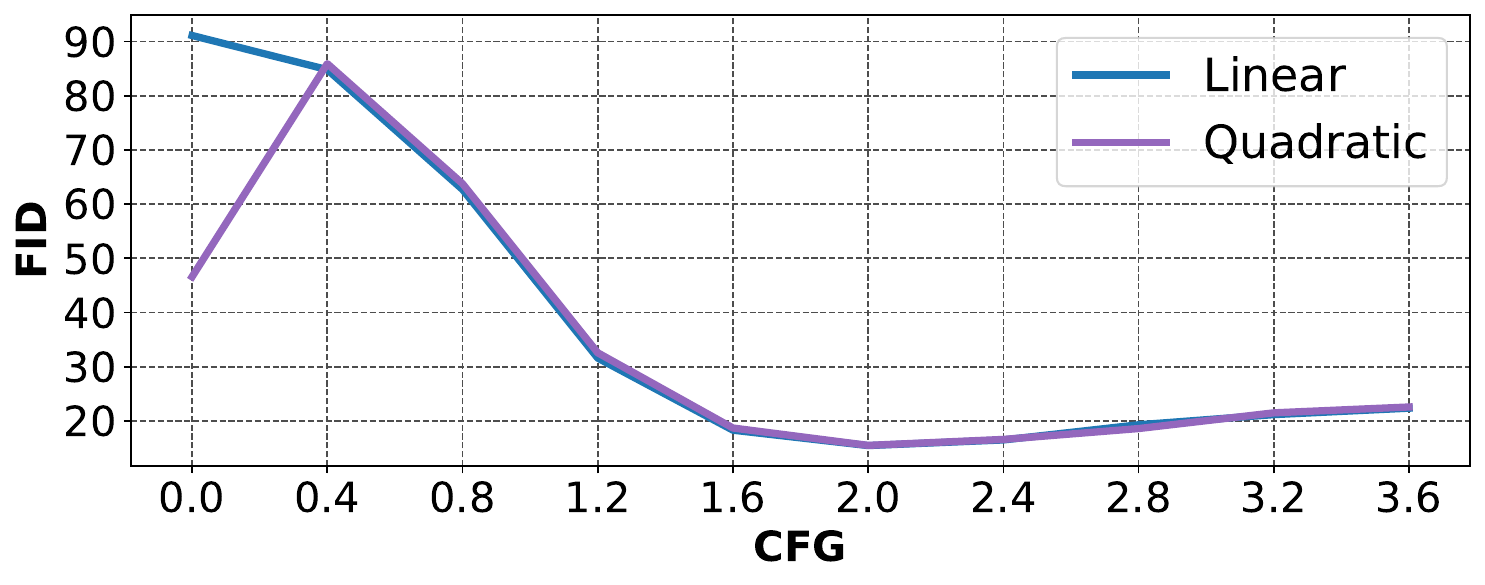}
        \caption{CFG in ITM.}
         \label{fig:itm_cfg_fid}
    \end{subfigure}

    \vspace{0.2cm} %

    \begin{subfigure}{0.33\textwidth}
      \includegraphics[width=\linewidth]{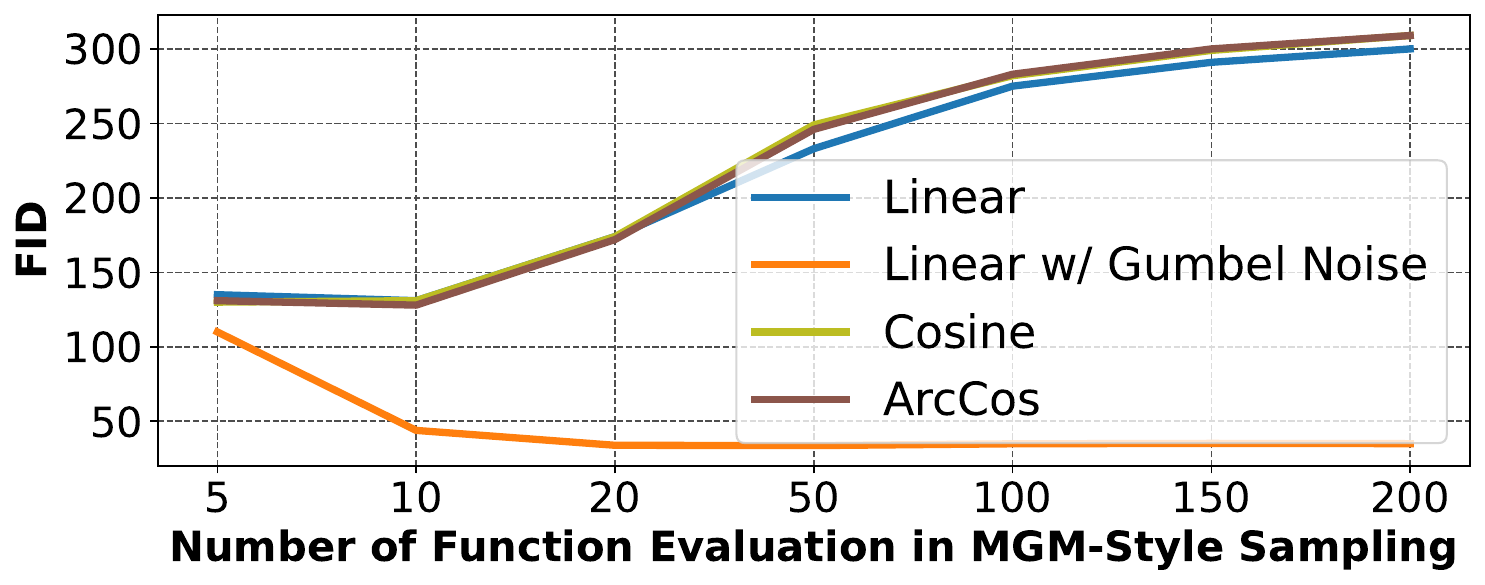}
        \caption{NFE in MGM-style sampling.}
         \label{fig:maskgit_nfe_fid}
    \end{subfigure}  
    \hfill 
    \begin{subfigure}{0.33\textwidth}
       \includegraphics[width=\linewidth]{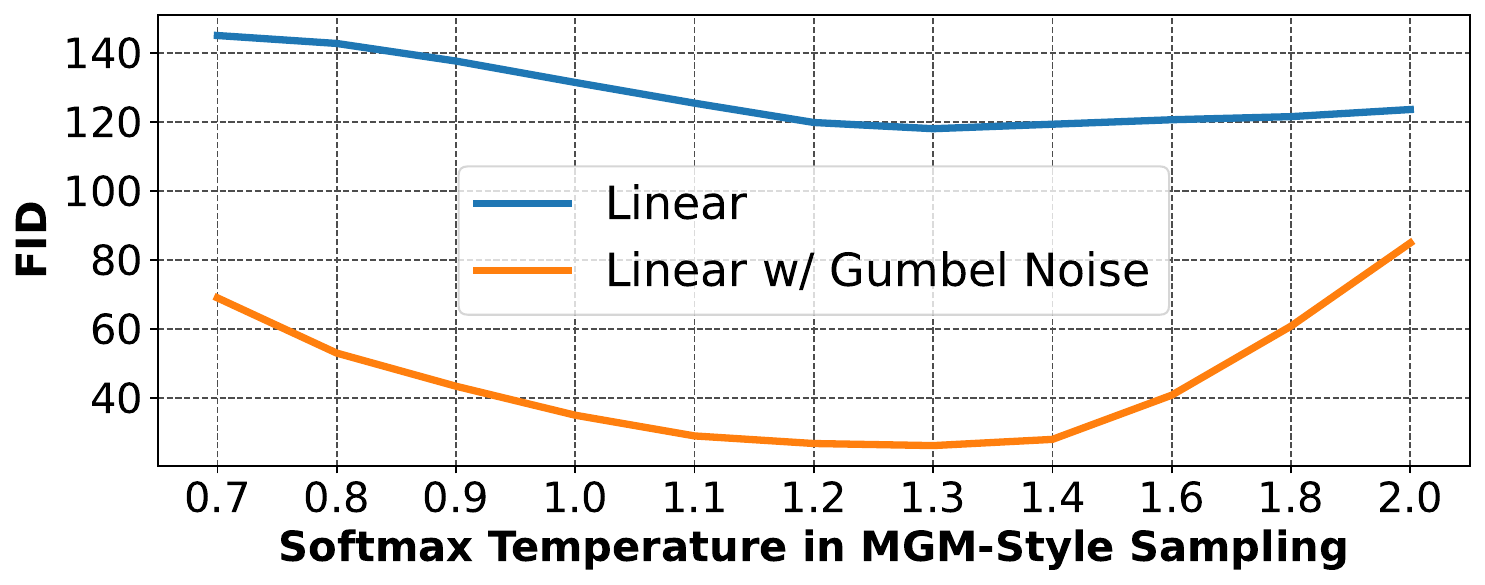}
        \caption{Temperature with 10 NFEs.}
        \label{fig:maskgit_temp_fid}
    \end{subfigure}
    \hfill
    \begin{subfigure}{0.33\textwidth}
        \includegraphics[width=\linewidth]{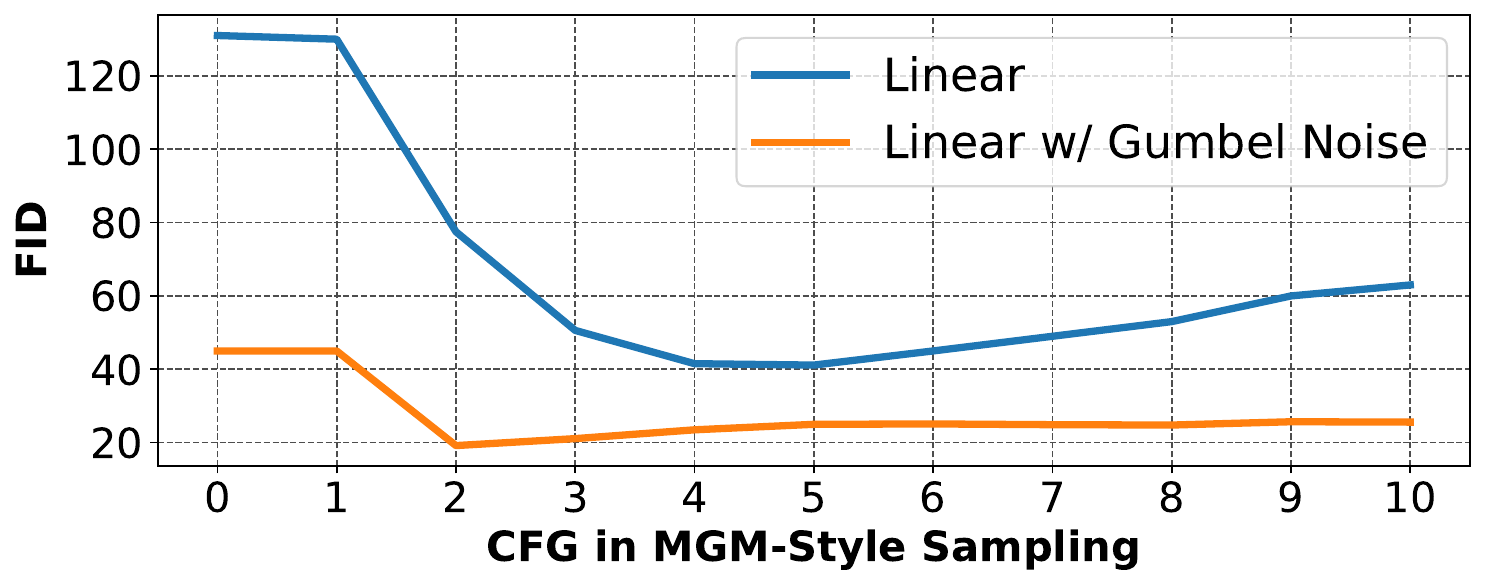}
        \caption{CFG with 10 NFEs.}
        \label{fig:maskgit_cfg_fid}
    \end{subfigure}

    \caption{\textbf{Ablation about  Explicit Timestep Model (ETM), Implicit Timestep Model (ITM), and Masked Generative Model(MGM) style Sampling} on ImageNet 256 dataset with FID-5k. All models are trained with linear schedulers by default. 
    }

    \label{fig:ablation_all}
\end{figure*}

\begin{figure}
    \centering
    \includegraphics[width=0.99\linewidth]{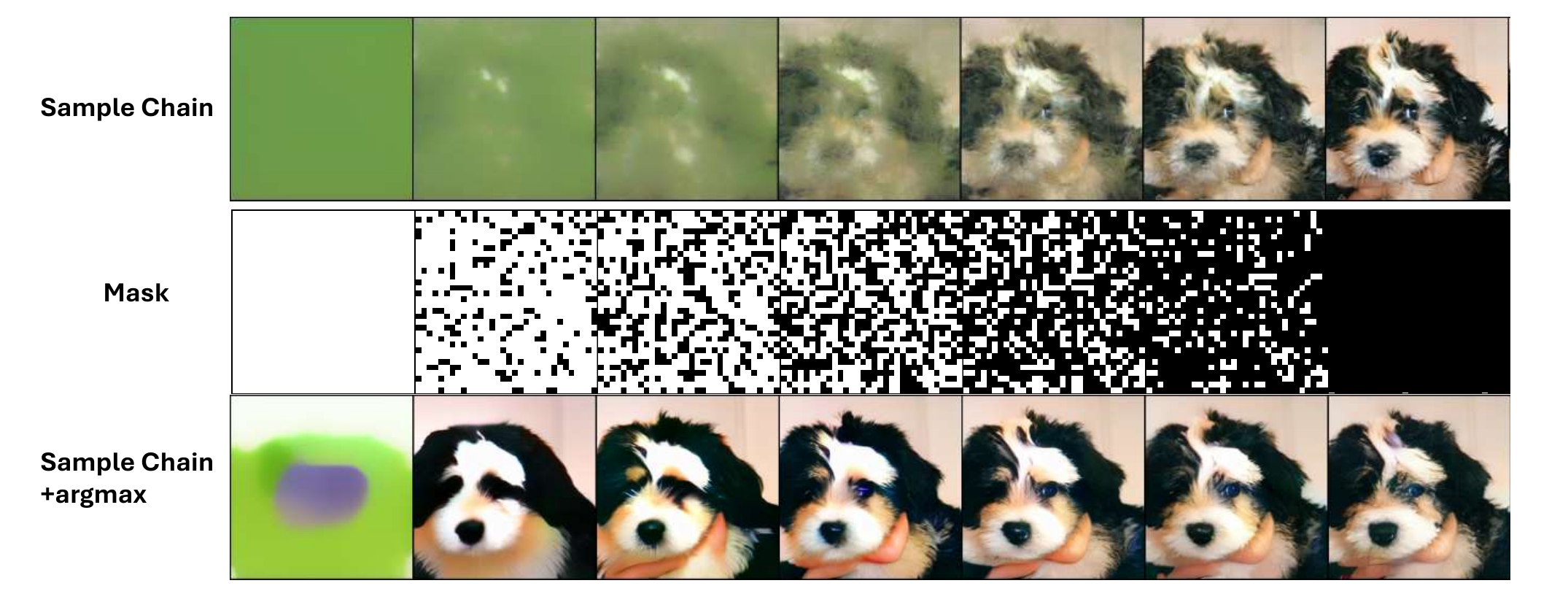}
    \caption{\textbf{Chain visualization} for ImageNet 256 with 100 timesteps with \texttt{argmax} applied. 
    }
    \label{fig:in256_chainvis}
\end{figure}

\subsection{Experimental Result}

\subsubsection{Main Result}

\paragraph{Image Generation.} We demonstrate our results on the COCO dataset, as shown in~\cref{tab:coco}. Our method achieves state-of-the-art performance compared to both continuous-state and discrete-state models. Notably, the Explicit Timestep Model and Implicit Timestep Models perform similarly.

We further showcase our performance on  ImageNet256 in ~\cref{tab:in256}. Compared to other conventional Autoregressive and Masked Image Model-based methods, our approach achieves competitive FID scores. The diffusion chain details can be seen in~\cref{fig:in256_chainvis}. 

For image generation, we consistently employ masking cross-entropy with $w(t)=1$. Empirically, we found that using the ELBO-derived weight $w(t)=\frac{\dot{\kappa}_t}{1-\kappa_t}$ yields suboptimal performance compared to $w(t)=1$. Additionally, we discovered that using cross-entropy without any masking leads to overfitting issues. For more details, please refer to the Appendix.

\paragraph{Video Generation.} To validate our method's scalability, we conduct experiments on the FaceForensics dataset, as shown in ~\cref{tab:video}. We adapt the conventional Latte~\cite{ma2024latte} models from continuous-state to discrete-state by inserting a learnable embedding for indices and a linear layer to map the logit dimension from $\mathbb{R}^{W\times H\times T}$ to $\mathbb{R}^{C\times W \times H \times T}$. As demonstrated in the table, our method performs better than its continuous-state counterpart under training settings, indicating that our approach scales from image to video generation. For further information, please refer to the Appendix.

\subsubsection{Ablation Study}

In this section, we analyze the design space of Masked Generative Models and Diffusion Models. Our ablation studies focus on key aspects of both models, including sampling timestep, softmax temperature, classifier-free guidance weight, and the impact of various sampling schedulers. Notably, our model demonstrates the ability to generalize to Masked Generative Model (MGM)-style sampling, further expanding its versatility.

For MGM-style models, we also explore a technique called linear Gumbel Noise. This involves linearly adding Gumbel noise to the confidence score in an annealing manner~\cite{chang2022maskgit,besnier2023MaskGit_pytorch}.

\paragraph{Number of Function Evaluation.} As shown in~\cref{fig:ablation_all}, the performance of Explicit and Implicit Timestep Models saturates around 100 steps. For MGM-style sampling without Gumbel noise, results are significantly worse across various noise schedules. However, after introducing Gumbel noise, performance saturates at just 10-20 steps. This rapid convergence occurs because MGM-style sampling relies on post-sampling confidence, potentially biasing optimal performance towards a lower number of function evaluations (NFE).

\paragraph{Implicit Timestep vs. Explicit Timestep.}
Comparing the first and second rows of~\cref{fig:ablation_all}, we observe nearly identical optimal FID for low NFE, optimal temperature, and optimal guidance strength of CFG. This similarity persists despite the different training approaches, suggesting that we can safely remove the timestep dependency when necessary.

\paragraph{The Influence of Scheduler. } From~\cref{fig:etm_nfe_fid,fig:itm_nfe_fid,fig:maskgit_nfe_fid}, we observe that the model's optimal FID converges to roughly the same value as we allow longer NFE. However, sampling with the training scheduler (linear) yields better performance. This indicates that misalignment indeed exists across various schedulers.

\paragraph{Softmax temperature.} As seen in~\cref{fig:itm_temp_fid,fig:etm_temp_fid,fig:maskgit_temp_fid}, the concept of softmax temperature originates from Masked Generative Models. We observe a sweet spot around 0.8, which interestingly remains consistent regardless of the scheduler choice or whether the timestep-dependence property is implicit or explicit. For MGM-style sampling, the temperature's sweet spot is approximately 1.2, becoming more pronounced after implementing Gumbel noise.

\begin{table}[]
    \centering
    \resizebox{0.98\linewidth}{!}{
    \begin{tabular}{l|c|cc}
    \toprule
    \textbf{Method} & \textbf{State} & \textbf{Frame-FID}$\downarrow$ & \textbf{FVD} $\downarrow$ \\
    \hline
     Latte~\cite{ma2024latte}& Continuous  & {21.20}   &  {99.53} \\ %
     \hline 
       Implicit Timestep Model(\textit{Our}) &  Discrete & {15.21}   & {81.20} \\
         \bottomrule
    \end{tabular}
     \vspace{-3mm}
    }
     \caption{\textbf{Video generation on FaceForensics dataset.}}
     \label{tab:video}
\end{table}

\begin{table}
    \centering
    \begin{tabular}{l|cc}
    \toprule
       \textbf{Method} & \textbf{FID(5k)}$\downarrow$ &  \textbf{mIOU} $\uparrow$  \\
         \hline
         Explicit Timestep Model & 34.4  & 89.1 \\
         Implicit Timestep Model & 33.8 & 90.1 \\
         \bottomrule
    \end{tabular}
   \vspace{-3mm}
        \caption{\textbf{Ablation results on Cityscapes.}
        }
    \label{tab:cs}
\end{table}

\paragraph{Strength of Classifier-free Guidance (CFG).} We also examine the impact of Classifier-free Guidance strength in~\cref{fig:itm_cfg_fid,fig:etm_cfg_fid,fig:maskgit_cfg_fid}. Interestingly, the optimal guidance strength remains consistent across different schedulers and timestep properties (implicit or explicit). For MGM-style sampling, we observe that the optimal point shifts to around 3, with the implementation of Gumbel noise yielding superior performance.

\paragraph{Churning last-step sampling by \texttt{argmax}.} To explore how close the sample is to the target distribution, we investigate a technique called \texttt{argmax}. This approach involves using a direct \texttt{argmax} operation on logit space instead of categorical sampling, effectively churning the sampling process with an extremely hard Dirac distribution. As shown in~\cref{fig:itm_nfe_fid,fig:etm_nfe_fid}, this technique significantly improves sampling performance at low NFE (number of function evaluations).

\vspace{-3mm}
\paragraph{Churning  entire sampling process by Top-p, Softmax temperature, CFG scale, Gumbel noise.}
Top-p (nucleus sampling) and softmax temperature control the overall randomness of the output. In contrast, guidance strength in Classifier-Free Guidance (CFG) steers the sampling process towards specific classes. Gumbel noise adds an extra layer of randomization to the ``confidence score" similar to MaskGiT~\cite{chang2022maskgit}). As shown in~\cref{fig:ablation_all}, these techniques can improve performance over the baseline, particularly when using fewer sampling steps (low NFE regimes). Due to space limit, we defer the experiment of top-p in Appendix.

\paragraph{Image-Segmask Pair Joint Training for Semantic Segmentation.} We conducted our primary experiments on the Cityscapes dataset, as shown in ~\cref{tab:cs}. Our generative models demonstrate versatility by performing both image-conditioned mask generation and mask-conditioned image generation. We evaluated these tasks using FID and mIOU metrics. The results reveal that our framework successfully handles both tasks with a single joint training process. Moreover, the comparable performance of the Explicit and Implicit timestep models suggests that we can leverage the \mm token to reframe discriminative tasks as an unmasking process.

\subsubsection{Visualization}

\paragraph{Chain visualization of ImageNet and Cityscapes.} To further validate the efficacy, we demonstrate the results of ImageNet256 generation in~\cref{fig:in256_chainvis}. We can already generate visually pleasing conditioned images, especially at low NFE, we also try to directly apply the \texttt{argmax} on the logit space, we can surprisingly obtain the image close to the final sampled images, which indicates that churning sampling can make the sampling more efficient.

\paragraph{\texttt{Argmax} alleviates scheduler misalignment and improves low-NFE performance.} As shown in ~\cref{fig:argmax_save_scheduler}, our model is trained with a linear scheduler. When we shift the sampling scheduler from linear to cosine, some tokens remain unmasked. Directly applying the \texttt{argmax} operation yields a reasonable result, compensating for this issue. This technique also improves performance when sampling at a low NFE, fully unmasking the remaining \mm tokens, as demonstrated in ~\cref{fig:argmax_save_scheduler}. Interestingly, when we apply \texttt{argmax} in the generative chains, it still produces samples 
 which are almost identical to the ground truth.

\section{Conclusion \& Future Works}

Our work Stochastic Interpolant extends discrete flow matching theory to vision tasks, generalizing from Explicit to Implicit Timestep Models. We analyze the intersection of Diffusion and Masked Generative Models, proposing dense-pixel prediction as an unmasking process. By integrating these elements, we achieve state-of-the-art performance on MS-COCO, competitive results on ImageNet 256, and demonstrate scalability to video datasets like Forensics. For future work, most mask-based methods can't remask tokens once unmasked, leading to irreversible denoising errors. CDCD~\cite{dieleman2022continuous_cdcd} addresses this, while ~\cite{liu2024think} proposes decoupling the process. Our method could potentially extend to these approaches, using discrete stochastic interpolants to address this limitation.

\section{Acknowledgment}

We would like to thank Timy Phan, Moyang Li for the extensive proofreading, and thanks Owen Vincent for technical support.
This project has been supported by the German Federal Ministry for Economic Affairs and Climate Action within the project “NXT GEN AI METHODS – Generative Methoden für Perzeption, Prädiktion und Planung”, the bidt project KLIMA-MEMES, Bayer AG, and the German Research Foundation (DFG) project 421703927. The authors gratefully acknowledge the Gauss Center for Supercomputing for providing compute through the NIC on JUWELS at JSC and the HPC resources supplied by the Erlangen National High Performance Computing Center (NHR@FAU funded by DFG).

{
    \small
    \bibliographystyle{ieeenat_fullname}
    \bibliography{main}
}

\clearpage
\appendix

\onecolumn %

\section{Appendix}
\tableofcontents 
\clearpage

\section{Potential Impact}

Currently, most mask-based methods share a common limitation: once a token is unmasked, it cannot be masked back again, which is also the main motivation of CDCD~\cite{dieleman2022continuous_cdcd}. This means denoising errors can't be reversed or corrected. ~\citet{liu2024think} propose a solution by decoupling the process into two separate parts, breaking this rule and thus fixing the accumulating error. We anticipate our method can be extended to their paradigm, the smoothing item in our discrete stochastic interpolants can be specifically design for this goal.

One other future work is to consider the modality-dependent masking schedule $\kappa_t$ in multi-modal learning. Our work can be seen as mean-parameterization since it leverages a prediction model for the mean of $x_0$ in a continuous state diffusion model, e.g., DDPM~\cite{ho2020denoising}. We anticipate similar to the case of continuous diffusion models, other parameterizations can still yield similar performance.

Despite the underlying conceptual similarities between Masked Generative Models and Diffusion Models, a substantial disparity in sampling efficiency persists between these methods. Discrete Flow Matching~\cite{campbell2024generative,gat2024discrete} typically requires approximately 2,048 sampling steps, whereas MaskGiT~\cite{chang2022maskgit} achieves comparable results with merely 18 steps. While the time-independence design in Discrete Flow Matching partially mitigates this gap, the difference in computational efficiency remains significant, highlighting an important area for future research and optimization. The sampling process can be quite flexible by location or probability, for example, we can unmask the token who is maximum probability named confidence score~\cite{chang2022maskgit}, or purity sampling in~\cite{tang2022improved_vqdiffusion,campbell2024generative}.

\section{Extra Discussions}

\subsection{Scheduler}

\paragraph{Smoothing factor.} Sometimes, the transition between the probability distribution $p(x_0)$ and $p(x_1)$ by such a binary interpolation of $\kappa_t,1-\kappa_t$ is too rigid, to alleviate this rigid issue, a smoothing factor $\gamma_t$ is introduced as:

\begin{equation}
\label{eq:discrete_stochastic_interpolants}
    p_{t|0,1}(x|x_0,x_1) = (1-\kappa_t) \delta_{x_0}(x) + p^{smooth}(x)+  \kappa_t \delta_{x_1}(x),
\end{equation}
where $\delta_{x_0}(x)$ can be instantialized as $\delta_{\mm}$ and  $\kappa_t \geq 0$. Intuitively, the smoothing term can bring two benefits: 1) data augmentation on the clean data. 2), it provides extra training for the corrector sampler.
  A common instantiation can be $\kappa_0=0$ ,$\kappa_1=1$,and $p^{smooth}(x)$ is a uniform distribution: $p^{smooth}(x) \sim U(0,1)$.  To sample from $p^{smooth}_{t|0,1}(x|x_0,x_1)$, the sample $x_t$ can be obtained by  from $ \sqrt{s\cdot \kappa_t \cdot (1-\kappa_t)}\mathbb{I}$, where $\mathbb{I}$ is an $\mathbb{R}^{L\times K}$ tensor full of one, s is used to control the stochasticity in the sampling process. The insight of the adding such a smoothing factor, is that, it encourages the interpolants between other uninvolved ``word" in the $k$-sized vocabulary, it's similar to the Brownian Bridge instantiation~\cite{tong2023simulation,albergo2023stochastic}. If $s=0$, it will degenerate to simple discrete interpolants of~\cref{eq:simple_interpolants}, if $s \rightarrow +\infty$, then the interpolants is over-smoothed, so that we are nearly sampling from a uniform distribution, which is not desired. Another interesting interpretation is to see it as data smoothing, with as a counterpart of label smoothing or data mixup~\cite{mixup,chen2020pointmixup}. 
  
  Uniform noise~\cite{campbell2024generative}, is a special case of our discrete stochastic interpolants. It injects non-\mm tokens to noise the real token, similar to the goal of our smoothing factor.  This design offers two benefits: it acts as a form of data augmentation and improves training for corrector sampling, allowing for the possibility of masking back data tokens—a topic beyond the scope of our paper. %

This Conditional Coupling is designed to encourage the network to better unmask the data when encountering a specific ratio of masked data. As previously mentioned, discrete data inherently contains timestep information, potentially causing misalignment between the mask ratio and the timestep. Fortunately, the network can be designed in an implicit timestep manner. By utilizing this time-independence design, the benefits of the conditional coupling scheduler can be better leveraged.

\paragraph{Why do we need various schedulers?} 
Assuming we have L-dimensional data, the number of possible masking cases is $C_{L}^{1}+C_{L}^{2}+...+C_{L}^{L}$, which approaches infinity as L grows large. For example, in text, L represents the number of tokens, while in computer vision, it could be 1,024 tokens under the LDM tokenizer for a $256\times256$ image. During training, it's impossible to enumerate all possible cases sufficiently. Therefore, we need to design various schedulers to facilitate the learning process. This suggests that changing the schedulers in the sampling process from those used in training will typically decrease performance, as we show empirically in our experiments. However, since the overall learning paradigm remains identical—with the masking schedule being the only difference—we demonstrate that fine-tuning with target samplers using minimal tokens can achieve performance similar to that of source schedulers.

One extra benefit of the discrete stochastic interpolates, it enables better sampling performance, e.g., In MaskGiT, To prevent MaskGIT from making overly greedy selections, random noise is added to the confidence scores, with its magnitude annealed to zero following a linear schedule.

In this way, we can uniformly cast these discriminative tasks as an unmasking process in discrete-state modeling. Classification and semantic segmentation have possibility sets of $[1]^{K}$ and $[L]^{K}$ respectively. This recasting enables us to better utilize the discrete nature of segmentation and pixel-level classification tasks.

\subsection{Connections with Other Methods}

\paragraph{Connection between BERT~\cite{kenton2019bert} and MAE~\cite{he2022masked_mae}.} BERT conducts masked training on data, feeding the masked tokens into the encoder network, while MAE removes these tokens from the encoder input. Our method shares similarities with their masking operations but differs in several ways: 1). Our masking operation is based on the masking schedule $\kappa_t$, whereas theirs uses a fixed probability. 2). Our unmasking operation is conducted progressively, while BERT and MAE unmask all tokens at once. 3). Our masking operation extends to other possible modalities, enabling more flexible sampling. To better illustrate the connection, we represent their masking schedule using our notation in a timestep-independent manner:

\begin{equation}
\kappa_t = d, 1-\kappa_t = 1-d,
\end{equation}
where $\kappa_t$ is always timestep-indepdent, and $d$ is the fixed probability of masking \textit{real} data.

\section{Extra Related Works}

\subsection{Discrete and Continuous Representation} 

The debate between discrete and continuous representations in generative models, as explored in works like GIVT~\cite{tschannen2023givt} and MAR~\cite{li2024autoregressive_mar}, has highlighted their respective strengths. However, these paradigms are not mutually exclusive. For instance, increasing the token vocabulary (e.g., using a large codebook of 162K can yield similar benefits to continuous-valued tokens, as shown in GIVT~\cite{tschannen2023givt}. This suggests that discrete representations can bridge the gap to continuous approaches under certain conditions.

Discrete-valued tokens offer several compelling advantages: (1) compatibility with large language models (LLMs)\cite{xie2024showo}, (2) efficient compression for edge devices, (3) improved visual understanding\cite{ge2023making_seed}, and (4) enhanced robustness in vision tasks~\cite{mao2021discrete}. Recent advances, such as LlaMaGen~\cite{sun2024autoregressive_llamagen}, have demonstrated that discrete tokenizers can be competitive with continuous latent space representations. Prominent examples include SD VAE~\cite{rombach2022high_latentdiffusion_ldm}, SDXL VAE~\cite{podell2023sdxl}, and OpenAI's Consistency Decoder, all widely adopted in diffusion models. These developments indicate that discrete representations in image tokenizers are no longer a bottleneck for image reconstruction. However, latent diffusion models~\cite{rombach2022high_latentdiffusion_ldm} have established that continuous representations still hold an edge in certain scenarios.

Discrete-state generative models can be categorized into two main types: (1) next-token prediction~\cite{sun2024autoregressive_llamagen} and (2) next-set-of-token prediction~\cite{chang2022maskgit}. These models aim to: (1) determine the sequence in which tokens are predicted (unmasked) and (2) identify the token to predict at each position using a score such as purity~\cite{tang2022improved_vqdiffusion}, confidence~\cite{chang2022maskgit}, or pure sampling~\cite{gat2024discrete}.

In exploring the potential of discrete visual tokens, works such as VQ-GAN~\cite{esser2021taming_vqgan}, VQ-VAE~\cite{van2017neural_vqvae}, GSQ~\cite{wang2024scaling}, and MAGVIT-v2~\cite{yu2023language_magvit2} have laid a strong foundation. Building on these efforts, we further investigate discrete diffusion models by leveraging pretrained discrete tokenizers. Recent work, such as~\cite{hu2024flow}, also explores intermediate learned embeddings while relying on continuous-state-based theories.

\subsection{Autoregressive and Non-autoregressive Models}

The concept of masking aligns with several other works \cite{austin2021structured_d3pm,gat2024discrete,ou2024absorbingdiscretediffusionsecretly_radd,campbell2024generative}. While these primarily focus on linear schedulers, our approach generalizes it to more flexible scheduling options.

Unified-IO~\cite{lu2022unifiedio} shares our goal of modeling joint distributions across multiple modalities. However, their approach applies a raw transformer to attend to different modalities without considering masked diffusion models. Unified-IO represents all image-like modalities using a pre-trained RGB VQ-GAN~\cite{esser2021taming_vqgan}, whereas we adopt modality-specific tokenizers tailored to each modality. Similarly, ImageBart~\cite{esser2021imagebart} amortizes the effort of handling different timesteps into separate models. In contrast, we consolidate this effort into a single model, avoiding the complexity of designing separate weights for each timestep.

On the other side,  there is a lot of exploration between the combination of Autoregressive and Non-Autoregressive models, e.g., Show-o~\cite{xie2024showo}, Transfusion~\cite{zhou2024transfusion}.
Show-o rephrases the problem from the perspective of MaskGit~\cite{chang2022maskgit}, and they don't have any concept about timestep in multi-stage training, but we aim to solve the problem from the perspective of discrete interpolants on a single stage and explore it around various noise schedules. Additionally, we extend the scope to encompass image generation, segmentation, and video generation.
Transfusion~\cite{zhou2024transfusion}, considers mixed state with discrete representation from text, and continuous representation from images, with dual loss from language and diffusion losses.
Although discrete-state model closely connects the Masked Generative Models and Diffusion Models, and ~\citet{kilian2024computational} tries to analyze the computation across different methods, we aim to provide a systematic analysis of the unified design space between Masked Generative Models and Diffusion Models.

Chameleon~\cite{team2024chameleon} introduces a family of token-based mixed-modal models capable of both comprehending and generating images. This approach represents all modalities as discrete tokens and utilizes a unified transformer-based architecture. The model is trained from scratch in an end-to-end manner for autoregressive modeling of visual generation. Our approach differs from this method. 

For more details, we encourage readers to refer to the survey~\cite{xiong2024autoregressive}.

\subsection{Implicit and Explicit Timestep in Diffusion Models}

Diffusion Models and Feature Representation interact across various domains~\cite{fuest2024diffusion,fundel2024distillation,yu2024representation_repa}. In most scenarios, diffusion features are extracted in a timestep-dependent manner—either through averaging~\cite{fundel2024distillation} or heuristic search~\cite{hu2023guided}. We extend this concept to develop Implicit Timestep diffusion models that incorporate timestep dependence within the models for discrete states. This approach is intuitive since masked image tokens inherently contain timestep information (i.e., masking ratio). While timestep independence should be possible in continuous states, limited research exists in this area. We believe this is due to network architecture limitations in detecting subtle changes in continuous time steps-based corruption of the original images. Nevertheless, several studies~\cite{stracke2024cleandift} demonstrate that networks can incorporate timestep dependence through fine-tuning, suggesting promise for implicit timestep models.

\section{Extra Results}
\label{sec:more_results}

\begin{figure*}
    \centering
    \begin{subfigure}{0.49\textwidth}
        \includegraphics[width=\linewidth]{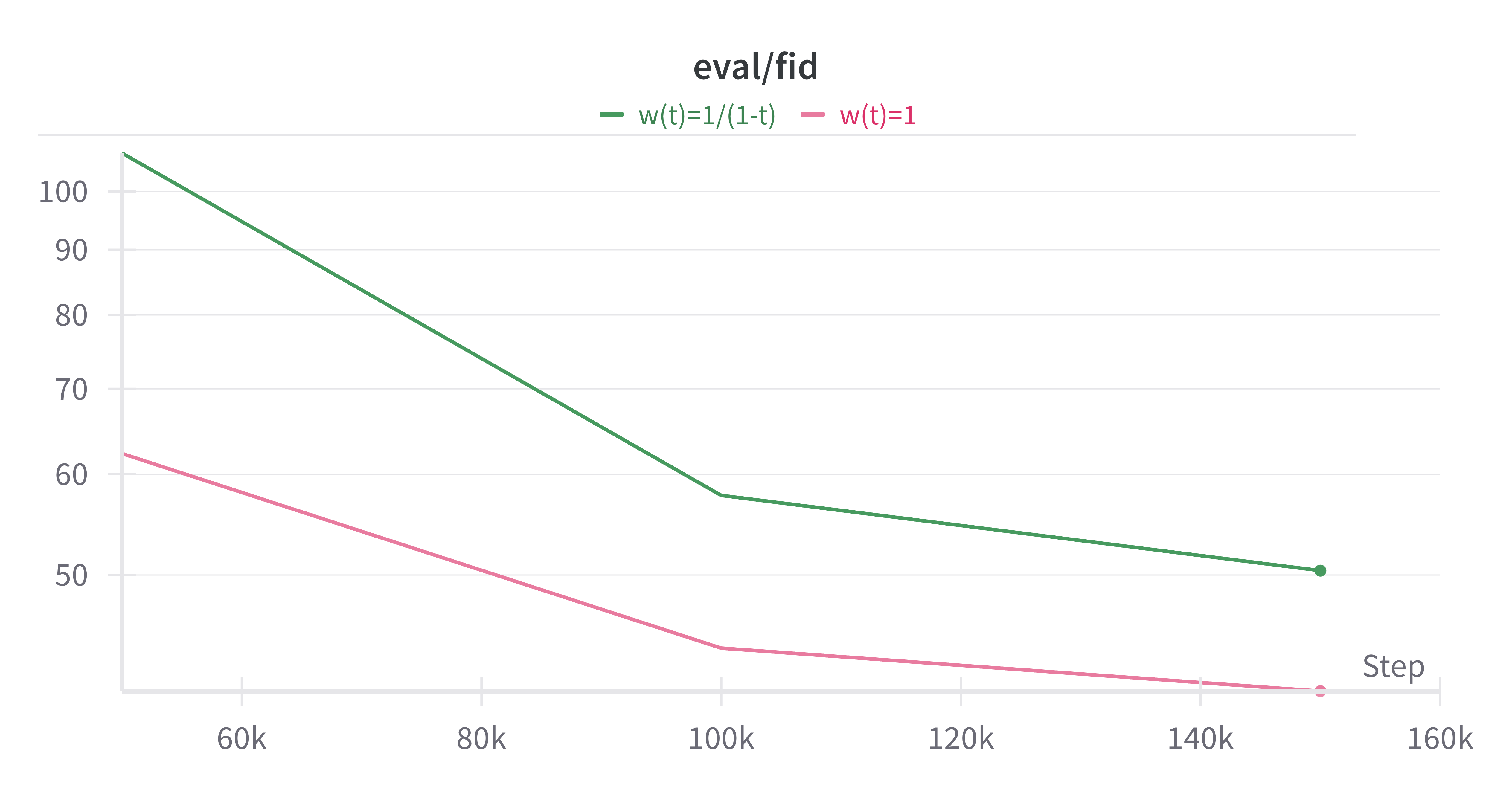}
        \caption{$w(t)=1$ is better than $w(t)=\frac{1}{1-t}$ for linear scheduler.}
        \label{fig:}
    \end{subfigure}
    \hfill
    \begin{subfigure}{0.49\textwidth}
        \includegraphics[width=\linewidth]{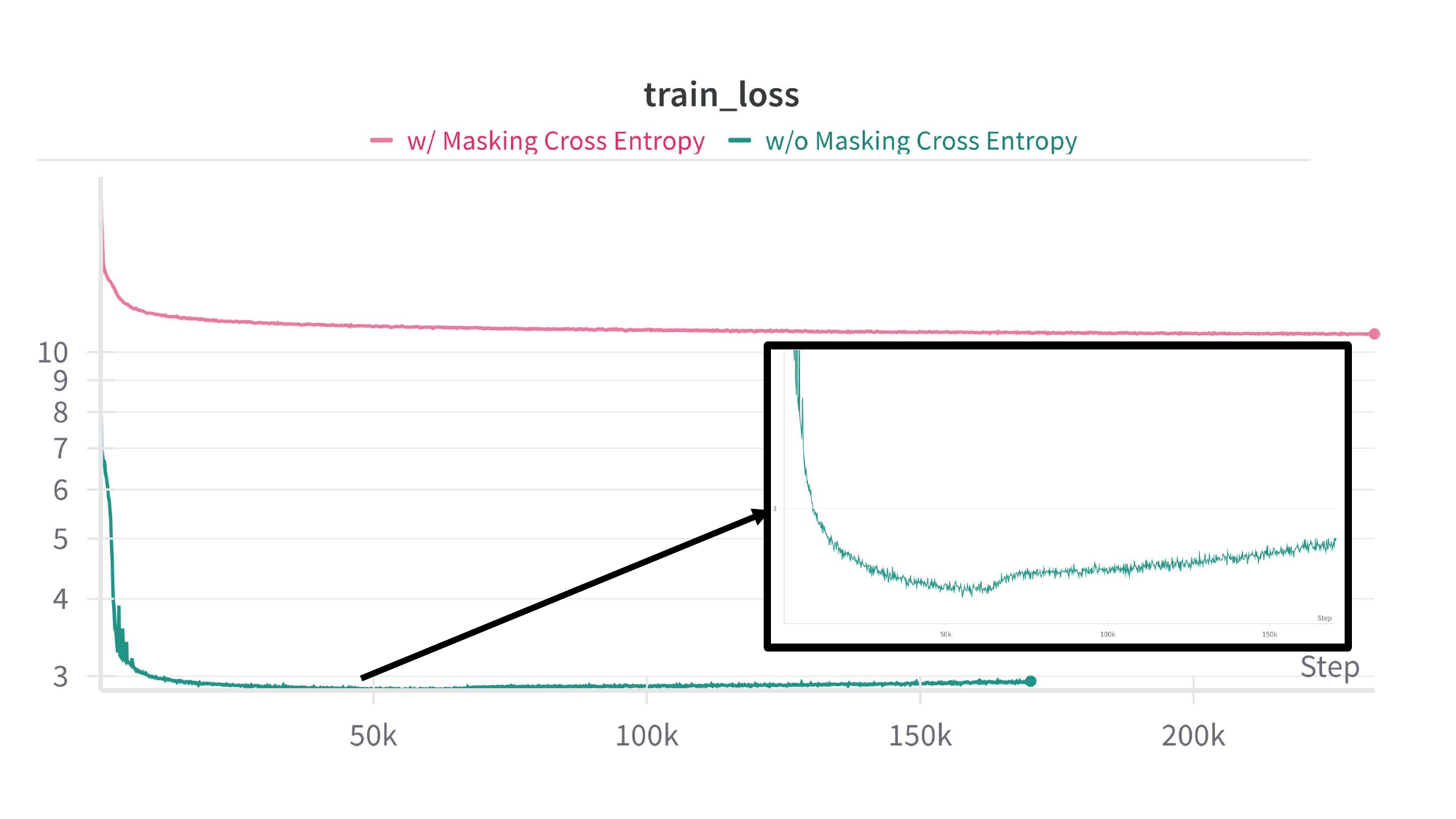}
        \caption{Masking on Cross-Entropy loss can avoid overfitting.}
        \label{}
    \end{subfigure}
    \caption{\textbf{Weighting and  Masking operation is important for training.} We train with batch size=1{,}024, and gradient clipping with 2.}
\label{fig:masking_and_weighting_is_critial}
\end{figure*}

Masking and suitable weighting on the cross-entropy loss are critical for performance, as demonstrated in~\cref{fig:masking_and_weighting_is_critial}.

\begin{figure*}
    \centering
    \begin{subfigure}{0.99\textwidth}
        \includegraphics[width=\linewidth]{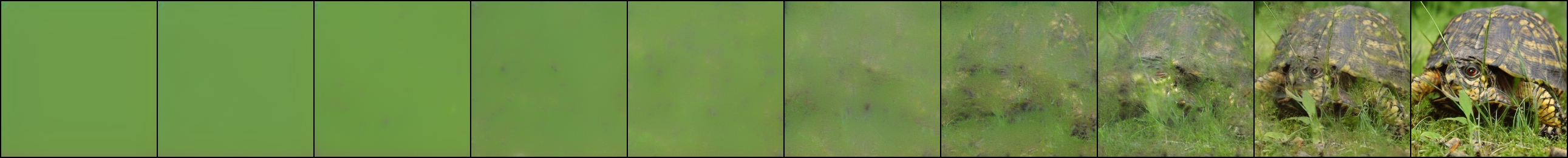}
        \caption{{Cosine Scheduler.}}
        \label{fig:cosine_forward}
    \end{subfigure}
    
    \vspace{0.2cm} %

        \begin{subfigure}{0.99\textwidth}
        \includegraphics[width=\linewidth]{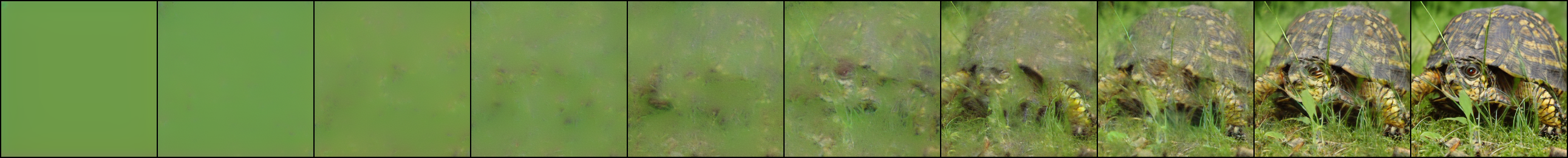}
        \caption{{Linear Scheduler.}}
        \label{fig:linear_forward}
    \end{subfigure}
    
    \vspace{0.2cm} %

        \begin{subfigure}{0.99\textwidth}
        \includegraphics[width=\linewidth]{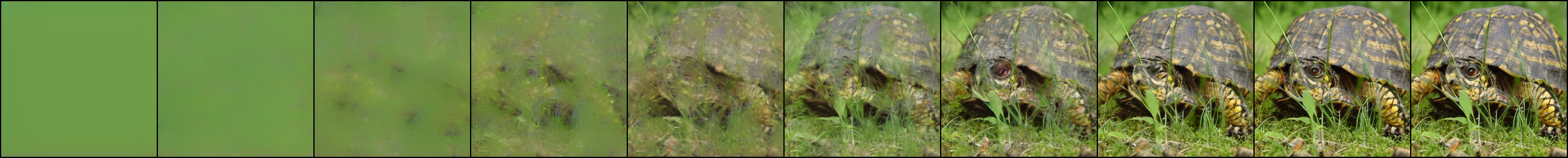}
        \caption{{Sine Scheduler.}}
        \label{fig:sine_forward}
    \end{subfigure}
    
    \vspace{0.2cm} %

    \caption{\textbf{Ablation about different schedulers, we mainly consider cosine, linear, and sine schedulers.}  
    }

    \label{fig:ablation_scheduler_forward_compare}
\end{figure*}

We visualize various schedulers in~\cref{fig:scheduler_vis}, with their corruption processes shown in~\cref{fig:ablation_scheduler_forward_compare}.

\begin{figure}
    \centering
    \begin{subfigure}[b]{0.49\textwidth}
        \centering
        \includegraphics[width=\textwidth]{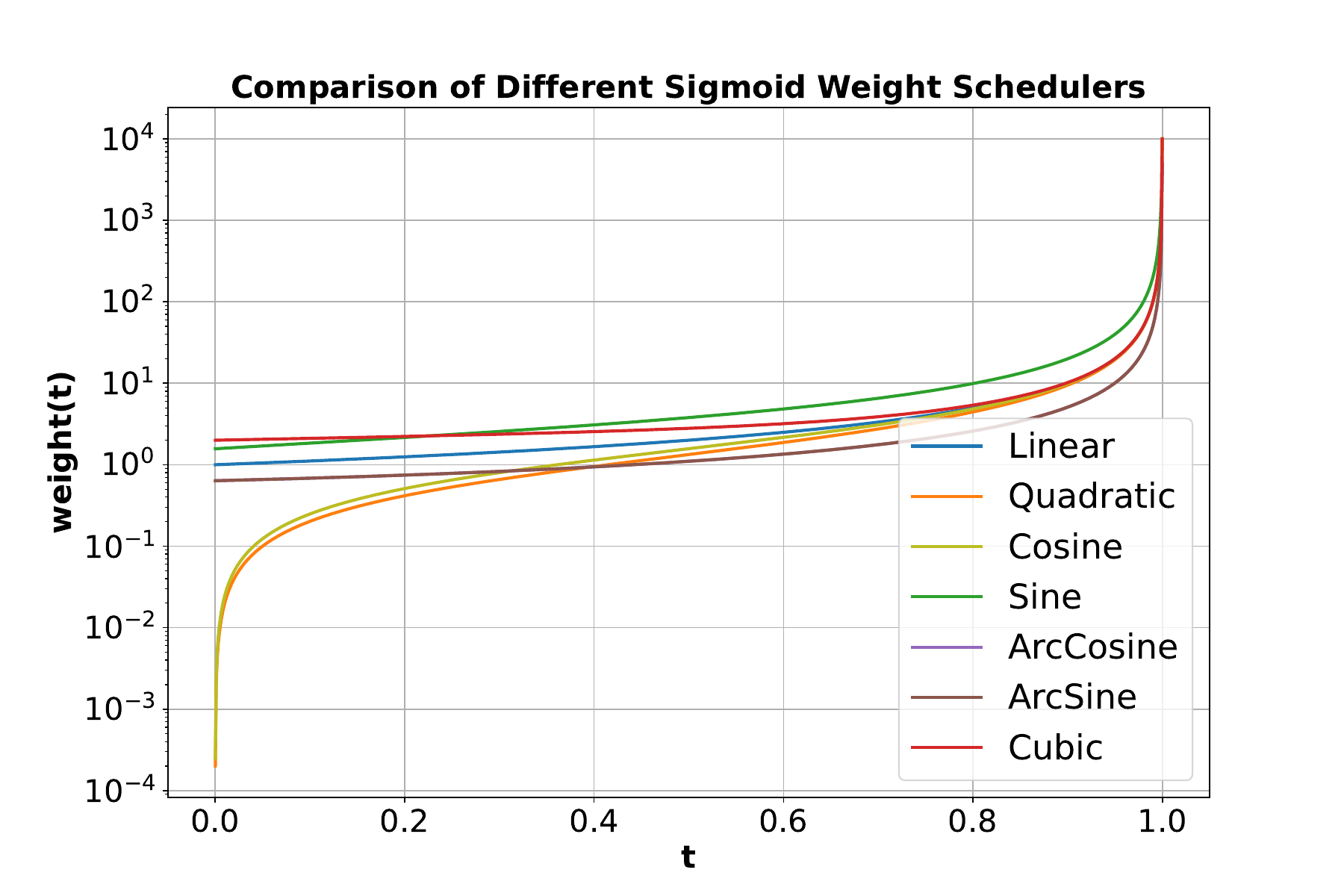}
        \caption{Weight: $w(t)$}
        \label{fig:figure1}
    \end{subfigure}
    \hfill
    \begin{subfigure}[b]{0.49\textwidth}
        \centering
        \includegraphics[width=\textwidth]{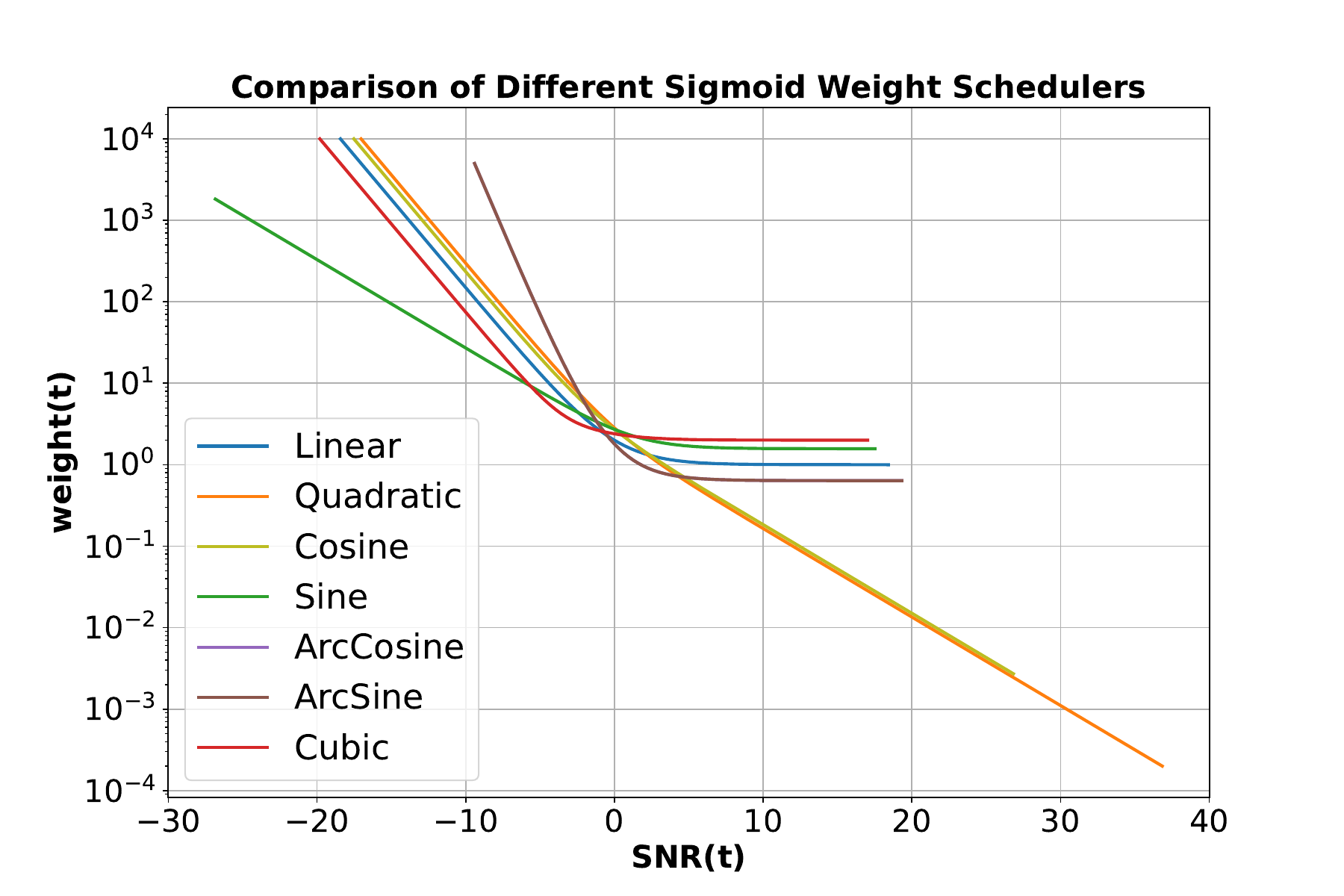}
        \caption{Weight \textit{v.s.} SNR(t)}
        \label{fig:figure2}
    \end{subfigure}
    \caption{\textbf{Weight $w(t)$ visualization.}}
    \label{fig:weight_wt_vis}
\end{figure}

For the weighting $w(t)$, we showcase its relationship with both time $t$ and signal-to-noise ratio $\text{SNR}(t)$ in~\cref{fig:weight_wt_vis}.

In~\cref{fig:ablation_all}, we provide visualizations comparing different CFG scales, softmax temperatures, Gumbel noise styles, and Gumbel noise temperatures.

\begin{figure}
    \centering
    \includegraphics[width=0.4\linewidth]{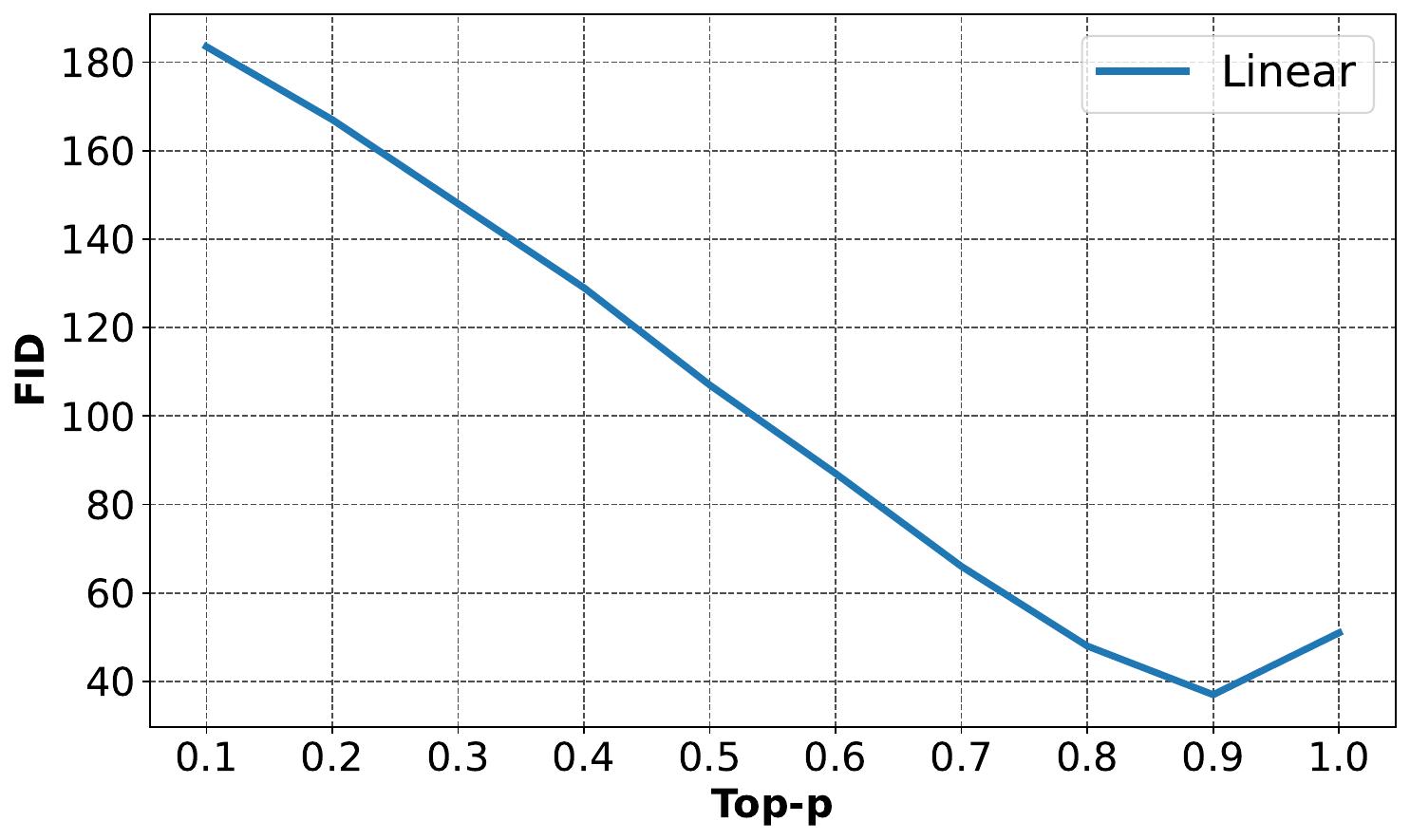}
        \caption{\textbf{Top-p v.s FID for ETM} for ImageNet.}
    \label{fig:ablate_in_topp}
\end{figure}

Our ablation study of top-p sampling on the ImageNet dataset in~\cref{fig:ablate_in_topp} reveals that top-p=0.9 yields optimal results.

\begin{figure*}[htbp]
    \centering
    \begin{subfigure}{0.33\textwidth}
        \includegraphics[width=\linewidth]{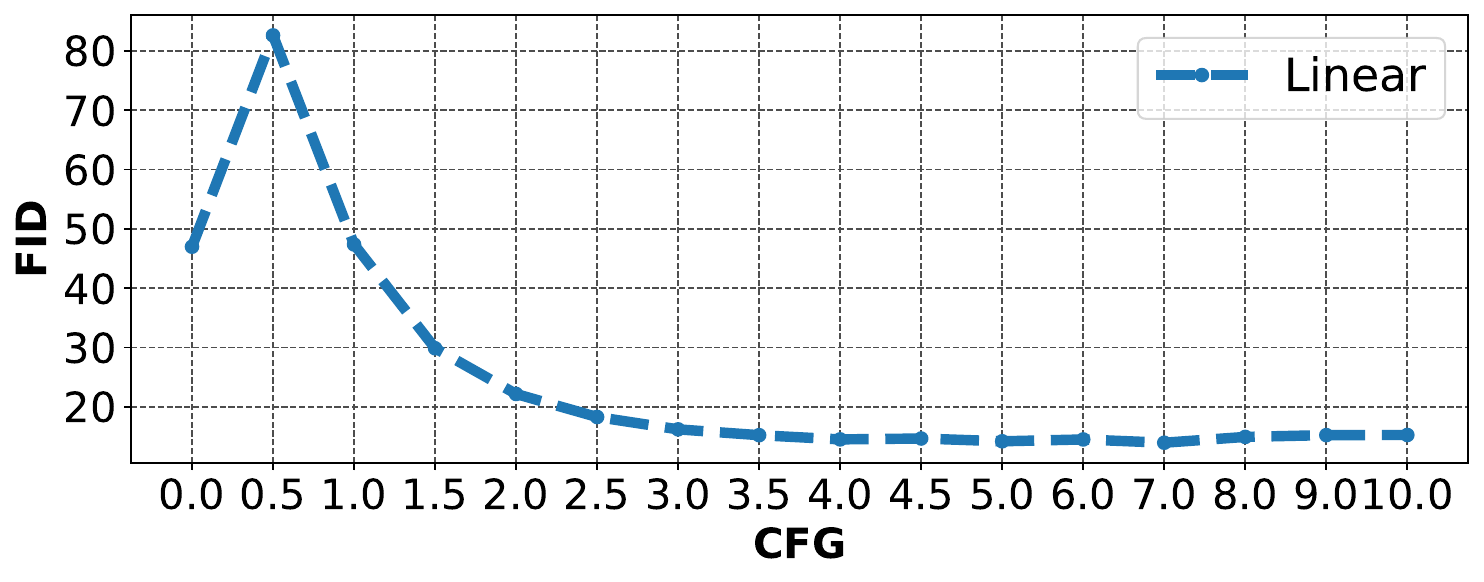}
        \caption{CFG v.s FID for ETM.}
        \label{}
    \end{subfigure}
    \hfill
    \begin{subfigure}{0.33\textwidth}
        \includegraphics[width=\linewidth]{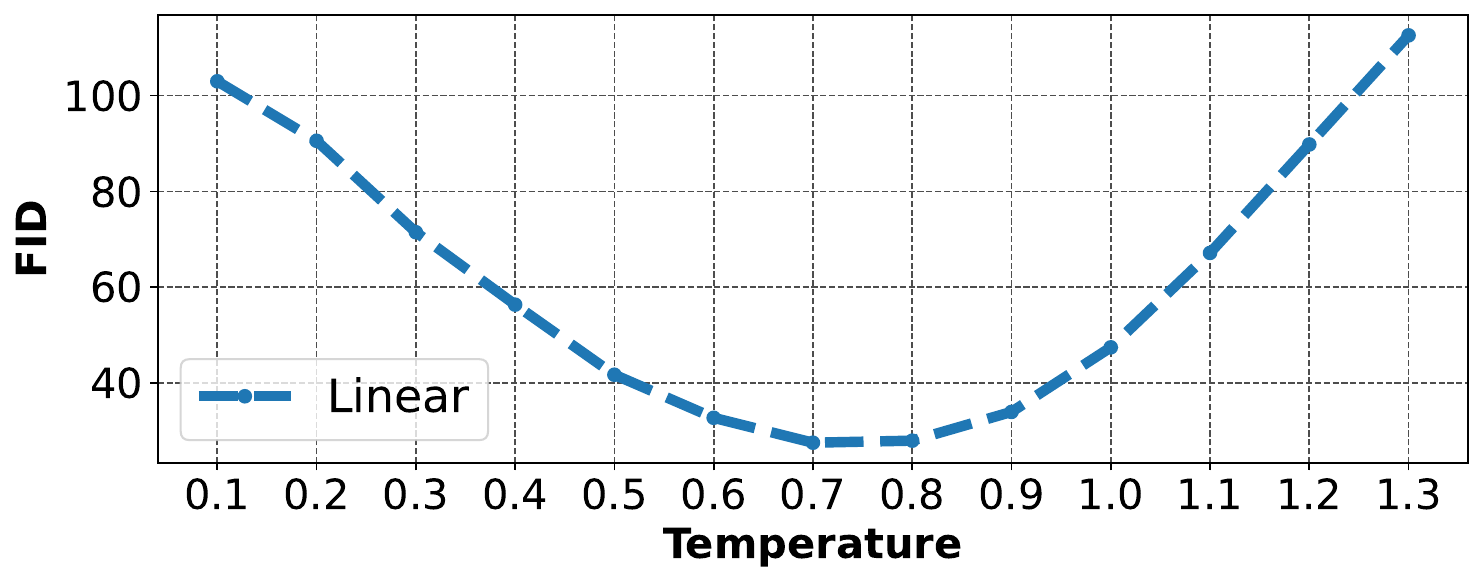}
        \caption{Temperature v.s FID  for ETM.}
        \label{}
    \end{subfigure}
    \hfill
    \begin{subfigure}{0.33\textwidth}
        \includegraphics[width=\linewidth]{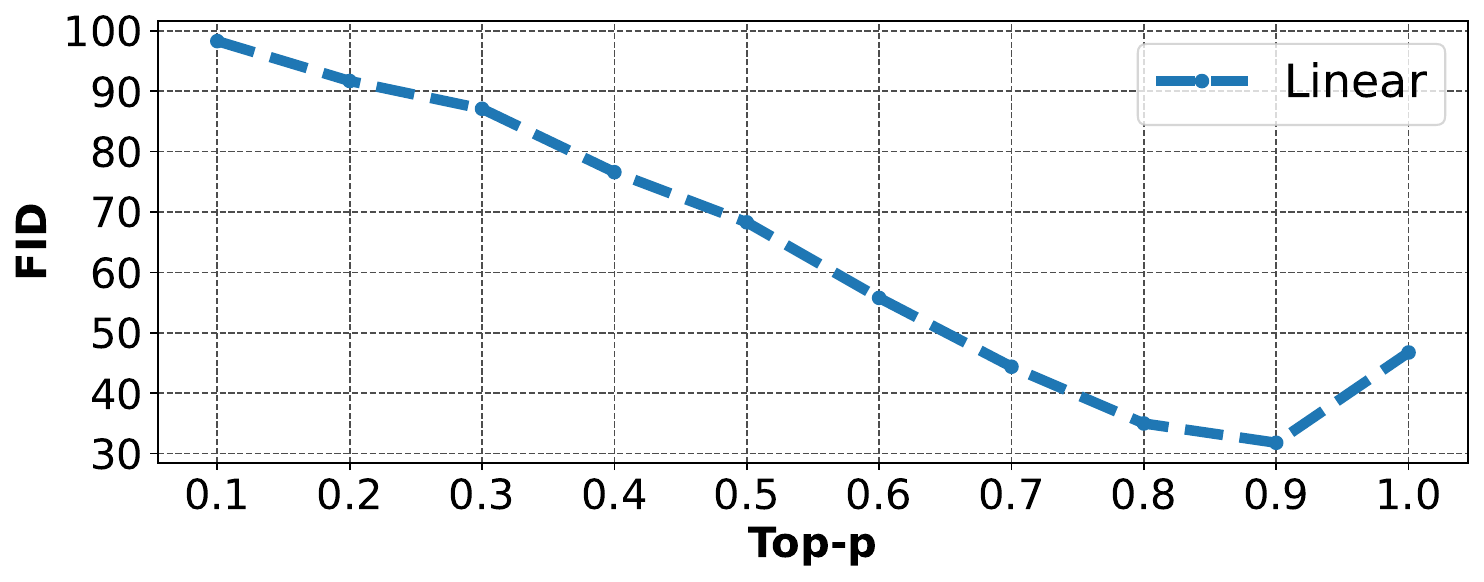}
        \caption{Top-p v.s FID  for ETM.}
        \label{}
    \end{subfigure}

    \vspace{0.2cm} %

    \caption{\textbf{The ablation of CFG,temperature,top-p of ITM(Implicit Timestep Model) in COCO dataset.}
    }

    \label{fig:coco_ablate}
\end{figure*}

We demonstrate different factors of the Implicit Timestep Model in~\cref{fig:coco_ablate}.

\begin{figure}
    \centering
    \includegraphics[width=0.89\linewidth]{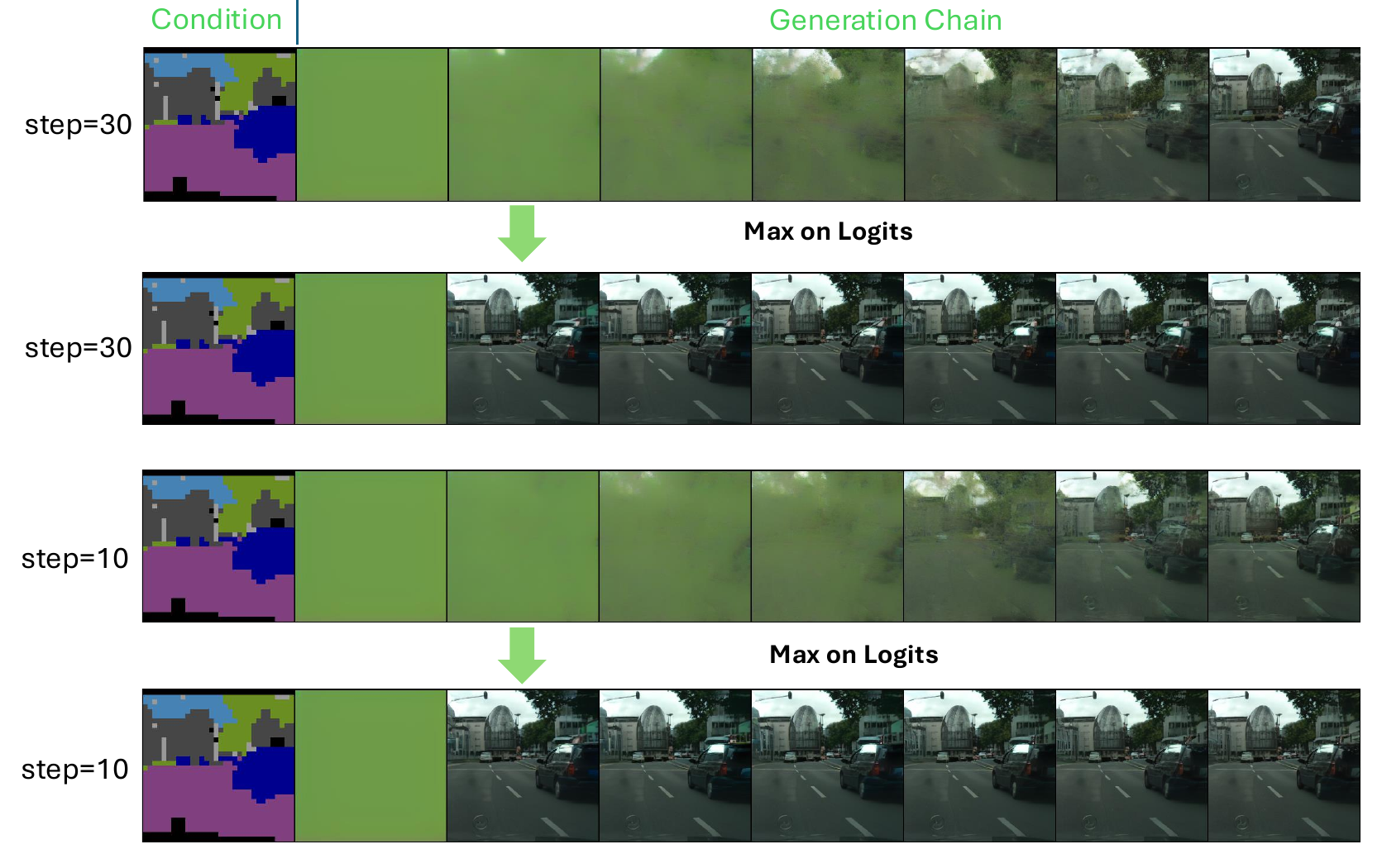}
    \caption{
    \textbf{Mask-conditioned image generation.}
    }
    \label{fig:mask_conditioned_img_gen}
\end{figure}

For Cityscapes dataset, we demonstrate segmentation mask-conditioned image generation in~\cref{fig:mask_conditioned_img_gen}, achieving visually pleasing results with relatively few function evaluations (NFE).

\begin{figure}
         \begin{subfigure}{0.99\textwidth}
        \includegraphics[width=\linewidth]{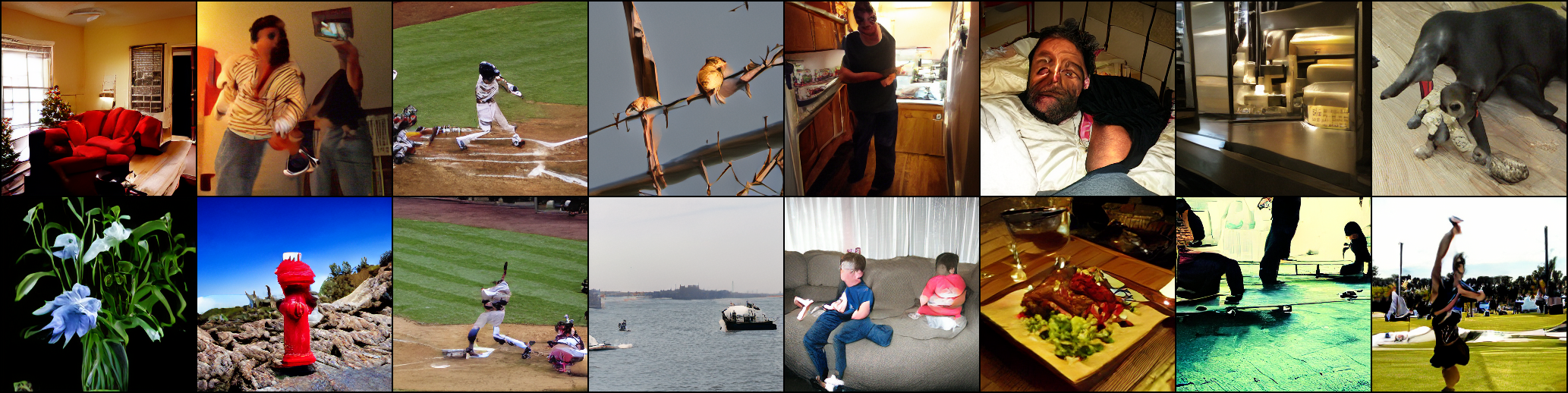}
    \end{subfigure}

 \vspace{0.2cm} %

      \begin{subfigure}{0.99\textwidth}
        \includegraphics[width=\linewidth]{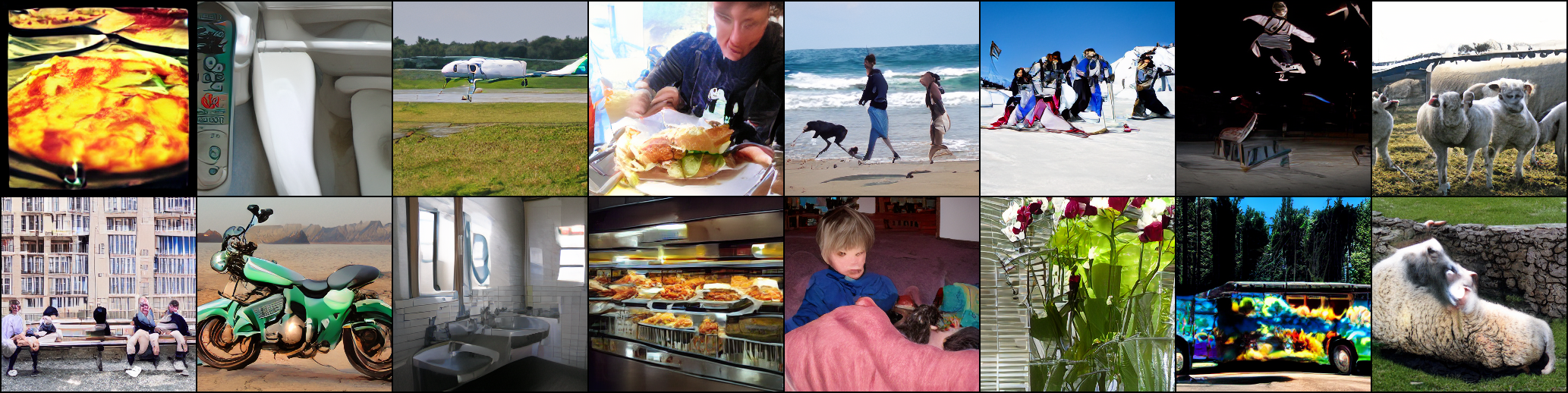}
    \end{subfigure}

    \caption{\textbf{Non cherry-picked visualization of MS COCO dataset.} CFG=4.5, FID-50k=5.8. }
    \label{fig:coco_vis}
\end{figure}

\begin{figure}
    
    \begin{subfigure}{0.99\textwidth}
        \includegraphics[width=\linewidth]{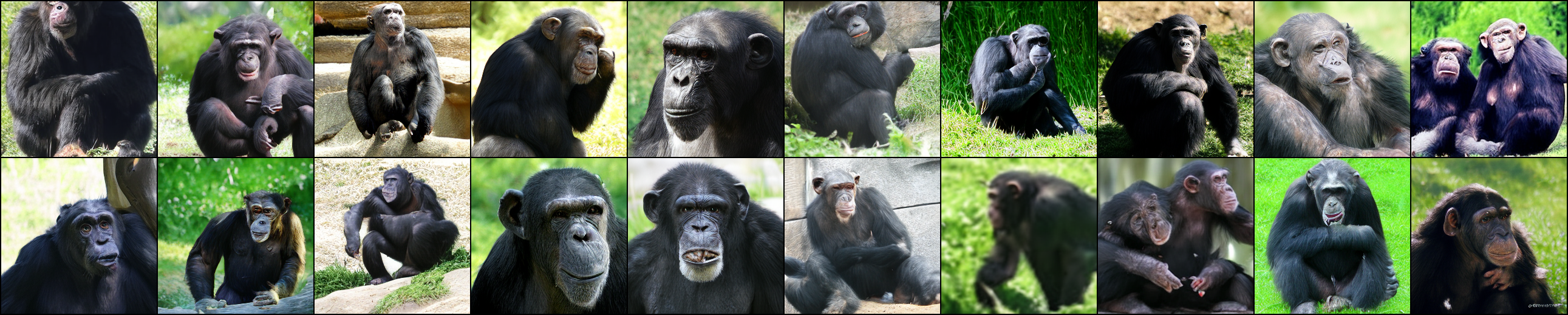}
    \end{subfigure}

 \vspace{0.2cm} %

    \begin{subfigure}{0.99\textwidth}
        \includegraphics[width=\linewidth]{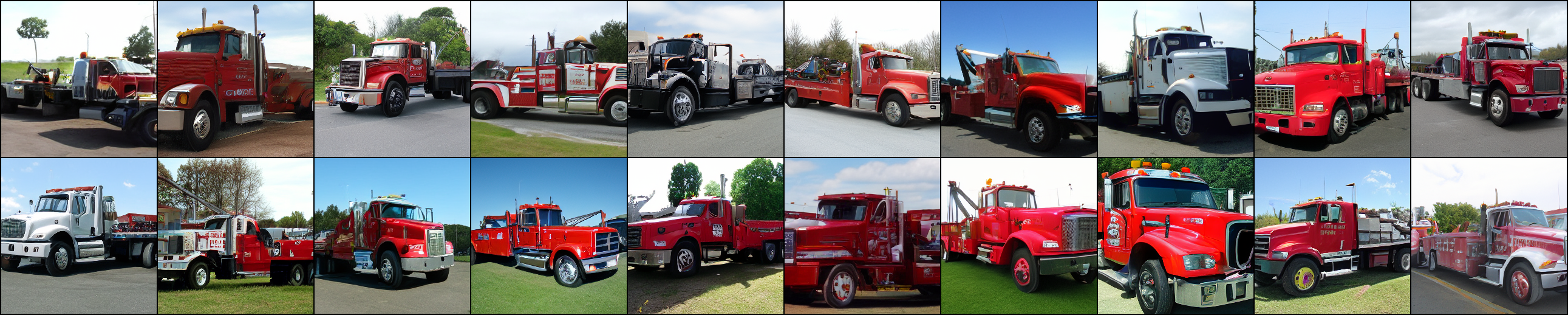}
    \end{subfigure}

 \vspace{0.2cm} %

     \begin{subfigure}{0.99\textwidth}
        \includegraphics[width=\linewidth]{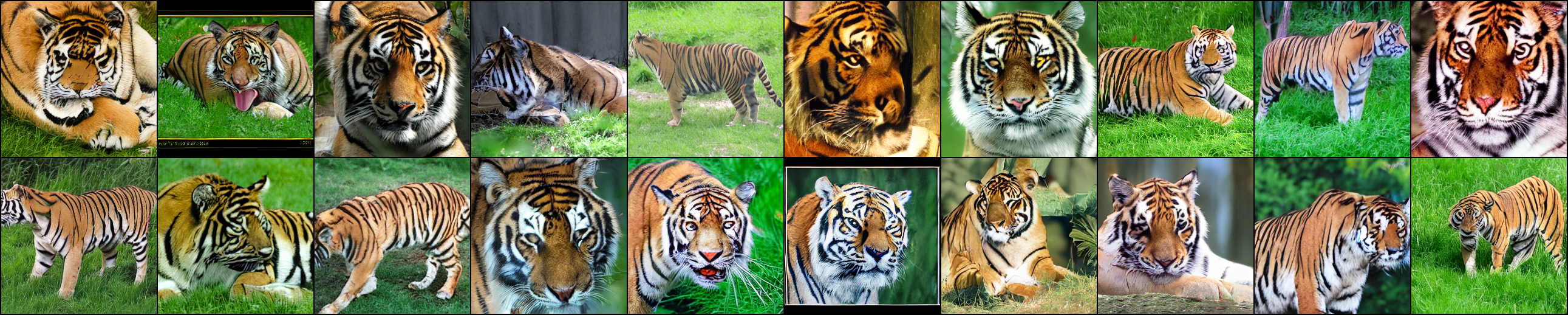}
    \end{subfigure}

 \vspace{0.2cm} %

     \begin{subfigure}{0.99\textwidth}
        \includegraphics[width=\linewidth]{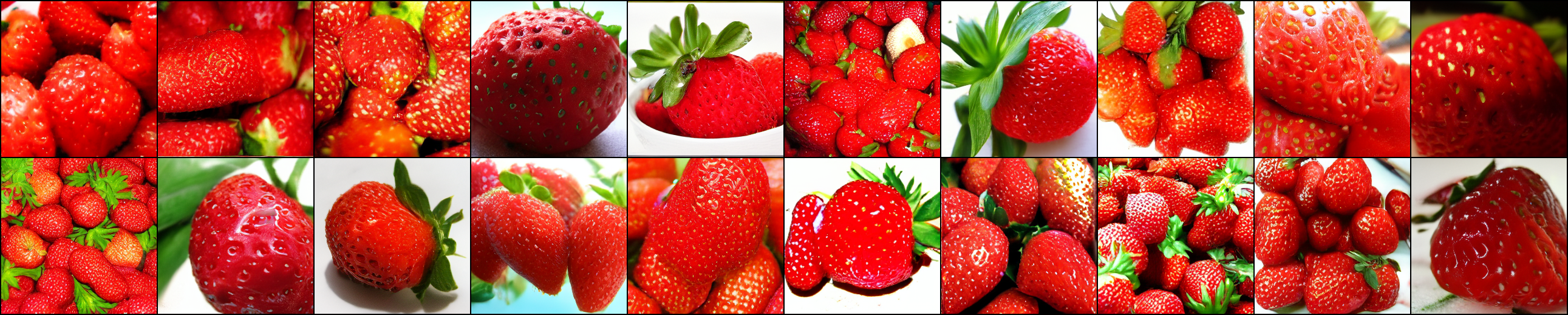}
    \end{subfigure}

\caption{\textbf{Non cherry-picked visualization of ImageNet256}. We sample for 20 steps with CFG=3.0, and temperature=1.3.
    }
    \label{fig:in256_vis}
\end{figure}

Sample visualizations from ImageNet and COCO datasets are shown in~\cref{fig:in256_vis} and ~\cref{fig:coco_vis} respectively.

\begin{figure}
    \centering
    \includegraphics[width=0.95\linewidth]{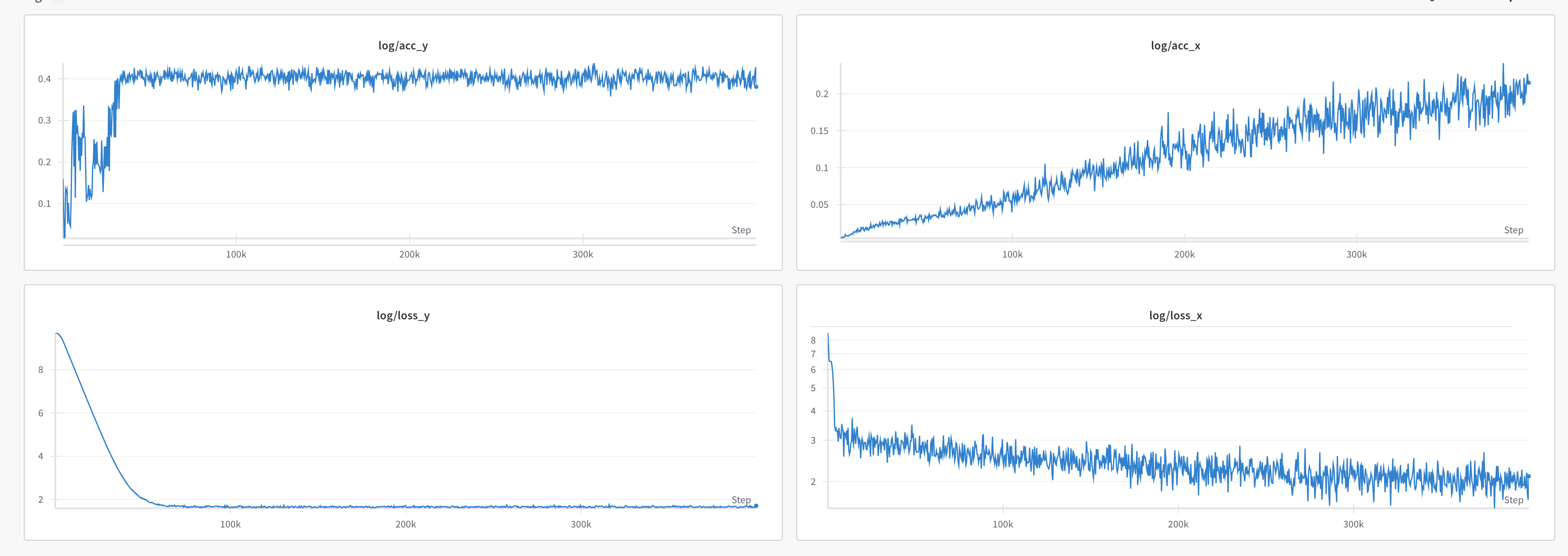}
    \caption{\textbf{The loss and accuracy trend of the joint Cityscapes training.}  x denotes the image, and y denotes the segmentation mask.} 
    
    \label{fig:cs_loss_acc_trend}
\end{figure}

For joint training on the Cityscapes dataset, we visualize discrete token prediction accuracy and loss in~\cref{fig:cs_loss_acc_trend}. The loss and accuracy patterns are similar between image and segmentation mask generation. However, the mask's cross-entropy loss shows greater stability than the image's loss. The higher accuracy for mask generation indicates it is an easier task than image generation.

\begin{figure*}
    \centering
    \begin{subfigure}{0.89\textwidth}
        \includegraphics[width=\linewidth]{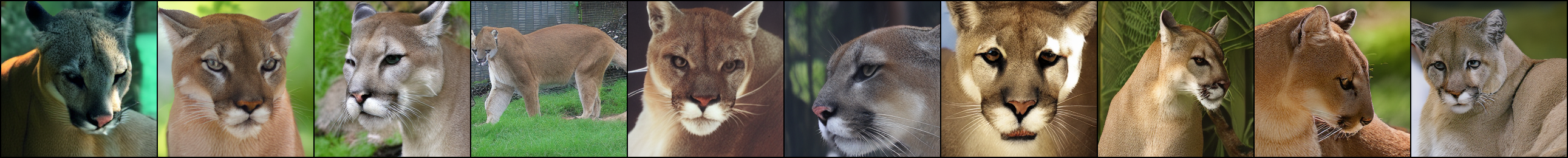}
        \caption{CFG=2, Temp=1.3, step20, gumbel=linear, gumbel\_temp=4.5.}
        \label{fig:}
    \end{subfigure}
    
    \vspace{0.2cm} %

    \begin{subfigure}{0.89\textwidth}
        \includegraphics[width=\linewidth]{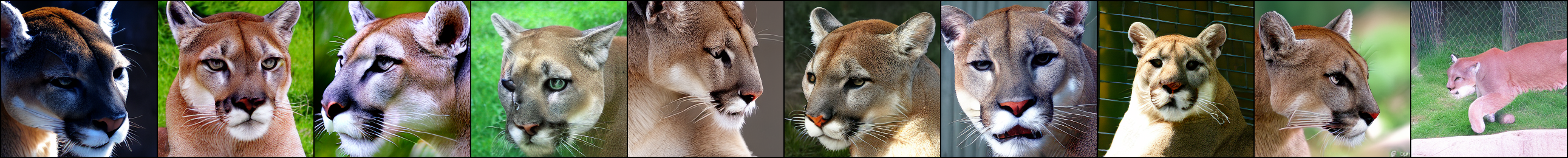}
        \caption{CFG=5, Temp=1.3, step20, gumbel=linear, gumbel\_temp=4.5.}
        \label{fig:}
    \end{subfigure}

     \vspace{0.2cm} %

         \begin{subfigure}{0.89\textwidth}
        \includegraphics[width=\linewidth]{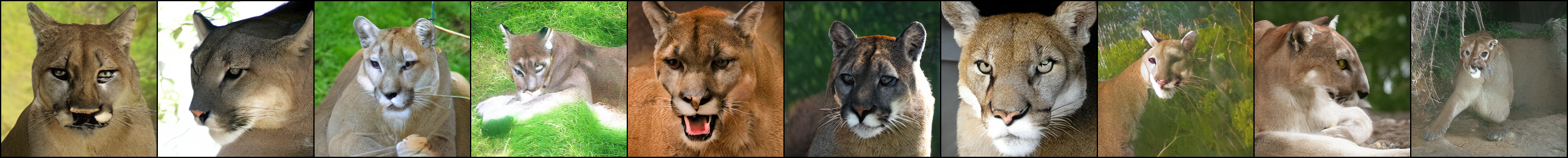}
        \caption{CFG=2, Temp=1.3, step10,gumbel=linear, gumbel\_temp=4.5.}
        \label{fig:}
    \end{subfigure}

      \vspace{0.2cm} %

        \begin{subfigure}{0.89\textwidth}
        \includegraphics[width=\linewidth]{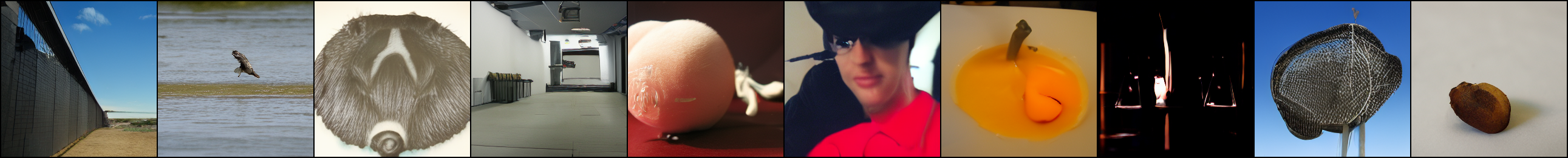}
        \caption{CFG=0, Temp=1.3, step20,gumbel=linear, gumbel\_temp=4.5.}
        \label{fig:}
    \end{subfigure}
    
    \vspace{0.2cm} %

        \begin{subfigure}{0.89\textwidth}
        \includegraphics[width=\linewidth]{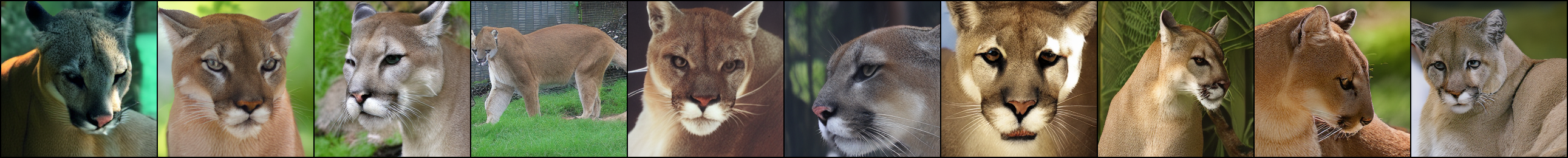}
        \caption{CFG=2, Temp=1.0, step20,gumbel=linear, gumbel\_temp=4.5.}
        \label{fig:}
    \end{subfigure}

    \vspace{0.2cm} %

    \begin{subfigure}{0.89\textwidth}
        \includegraphics[width=\linewidth]{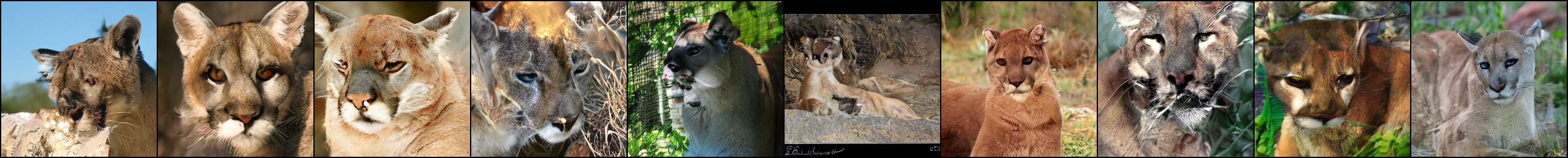}
        \caption{CFG=2, Temp=1.3, step20, gumbel=random, gumbel\_temp=4.5.}
        \label{fig:}
    \end{subfigure}

 \vspace{0.2cm} %

    \begin{subfigure}{0.89\textwidth}
        \includegraphics[width=\linewidth]{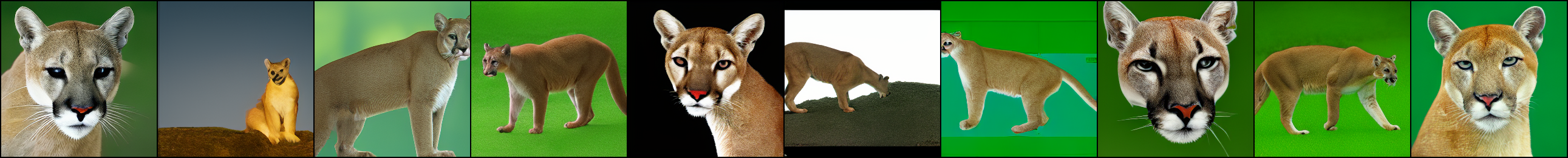}
        \caption{CFG=2, Temp=1.3, step20, gumbel=warmup first 2 steps, gumbel\_temp=4.5.}
        \label{fig:}
    \end{subfigure}

    \vspace{0.2cm} %

    \begin{subfigure}{0.89\textwidth}
        \includegraphics[width=\linewidth]{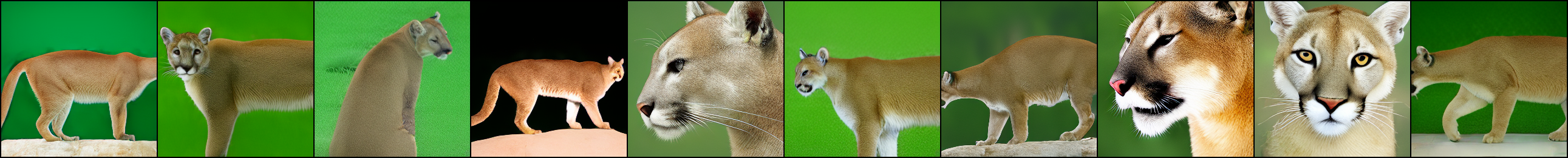}
        \caption{CFG=2, Temp=1.3, step20, gumbel=none, gumbel\_temp=4.5.}
        \label{fig:}
    \end{subfigure}

    \vspace{0.2cm} %

    \begin{subfigure}{0.89\textwidth}
        \includegraphics[width=\linewidth]{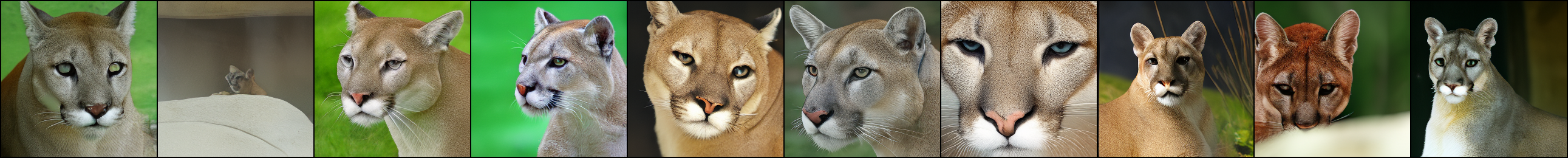}
        \caption{CFG=2, Temp=1.3, step20, gumbel=linear, gumbel\_temp=1.}
        \label{fig:}
    \end{subfigure}

 \vspace{0.2cm} %

    \begin{subfigure}{0.89\textwidth}
        \includegraphics[width=\linewidth]{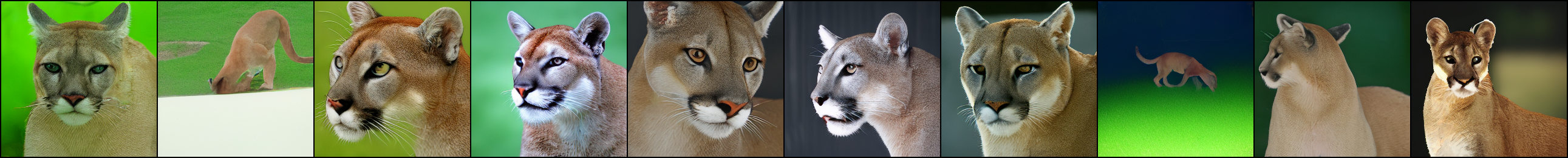}
        \caption{CFG=2, Temp=1.3, step20, gumbel=linear, gumbel\_temp=0.5.}
        \label{fig:}
    \end{subfigure}

    \caption{\textbf{Ablation about different sampling with the same class and same seed. 
    } We mainly compare with different CFG scales, softmax temperature, gumbel noise style, and the temperature of the gumbel noise.
    }

    \label{fig:ablation_all}
\end{figure*}

\begin{figure*}
    \centering
    \begin{subfigure}{0.49\textwidth}
        \includegraphics[width=\linewidth]{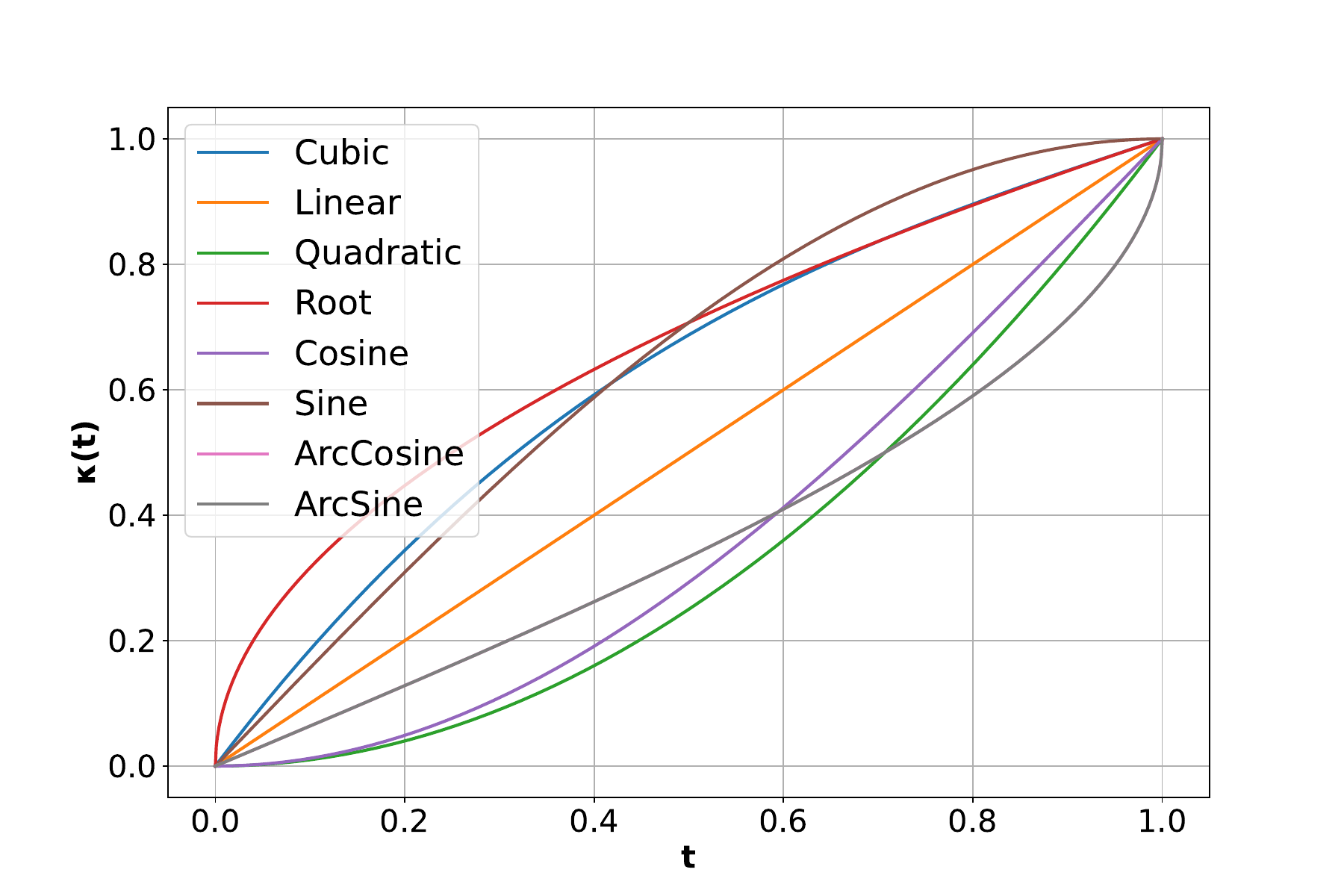}
    \caption{{Various schedulers $\kappa(t)$.}}
    \label{fig:scheduler_vis}
    \end{subfigure}
    \hfill
    \begin{subfigure}{0.49\textwidth}
        \includegraphics[width=\linewidth]{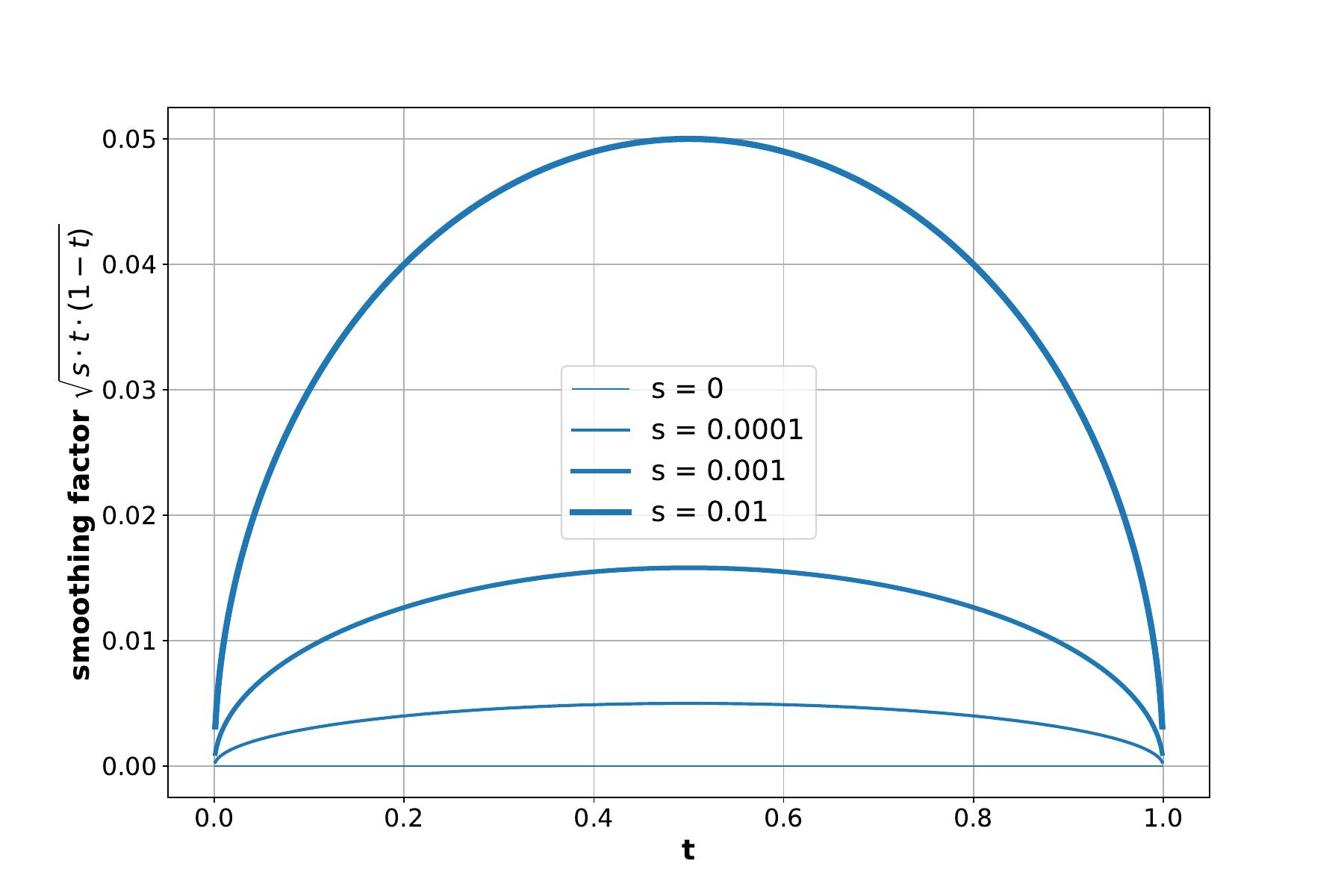}
    \caption{{Visualization of various smoothing factors.}}
    \label{fig:smoothing_factor_vis}
        \label{}
    \end{subfigure}

    \caption{\textbf{Scheduler Visualization.}}
    \label{fig:scheduler_vis}
\end{figure*}

\begin{figure}
    \centering
    \includegraphics[width=0.89\linewidth]{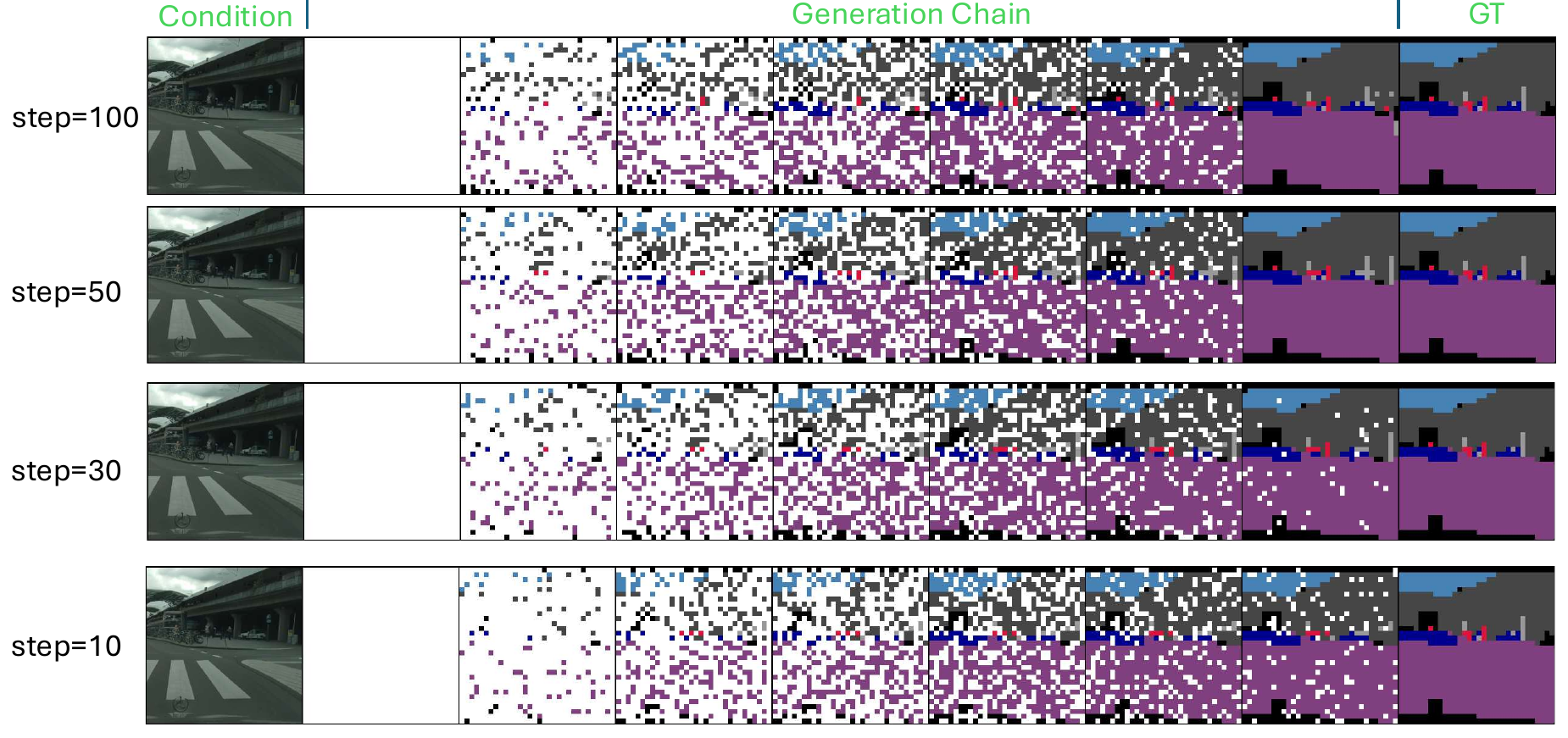}
    \caption{\textbf{Progressive chain visualization for different steps} in image-conditioned segmask generation on  Cityscapes datasets. We use CFG scale=3. }
    \label{fig:enter-label}
\end{figure}

\begin{figure}
    \centering
    \includegraphics[width=0.89\linewidth]{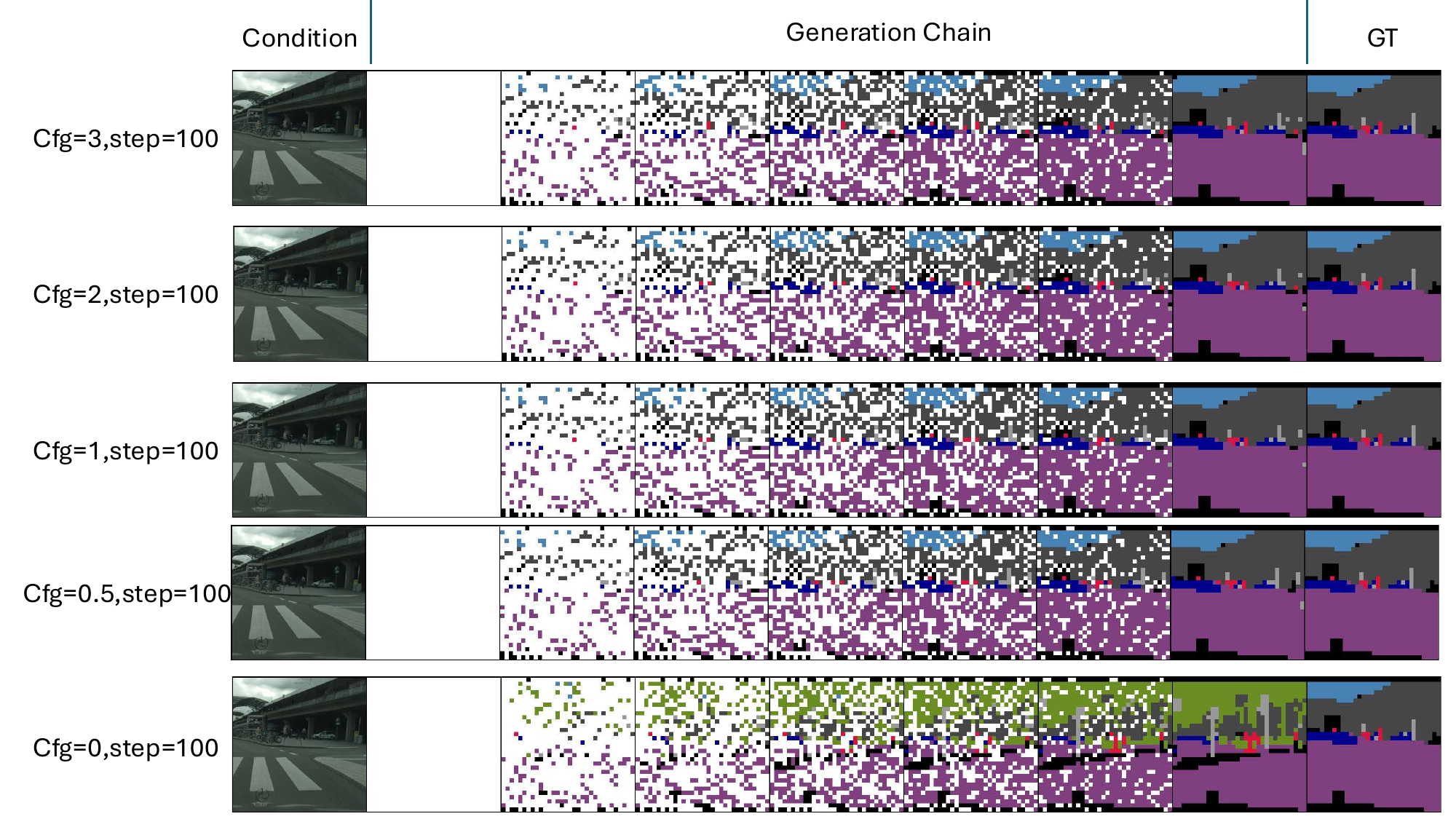}
    \caption{\textbf{The guidance scale of classifier-free guidance}  for image-conditioned segmask generation on Cityscapes.}
    \label{fig:enter-label}
\end{figure}

\begin{figure}
    \centering
    \includegraphics[width=0.79\linewidth]{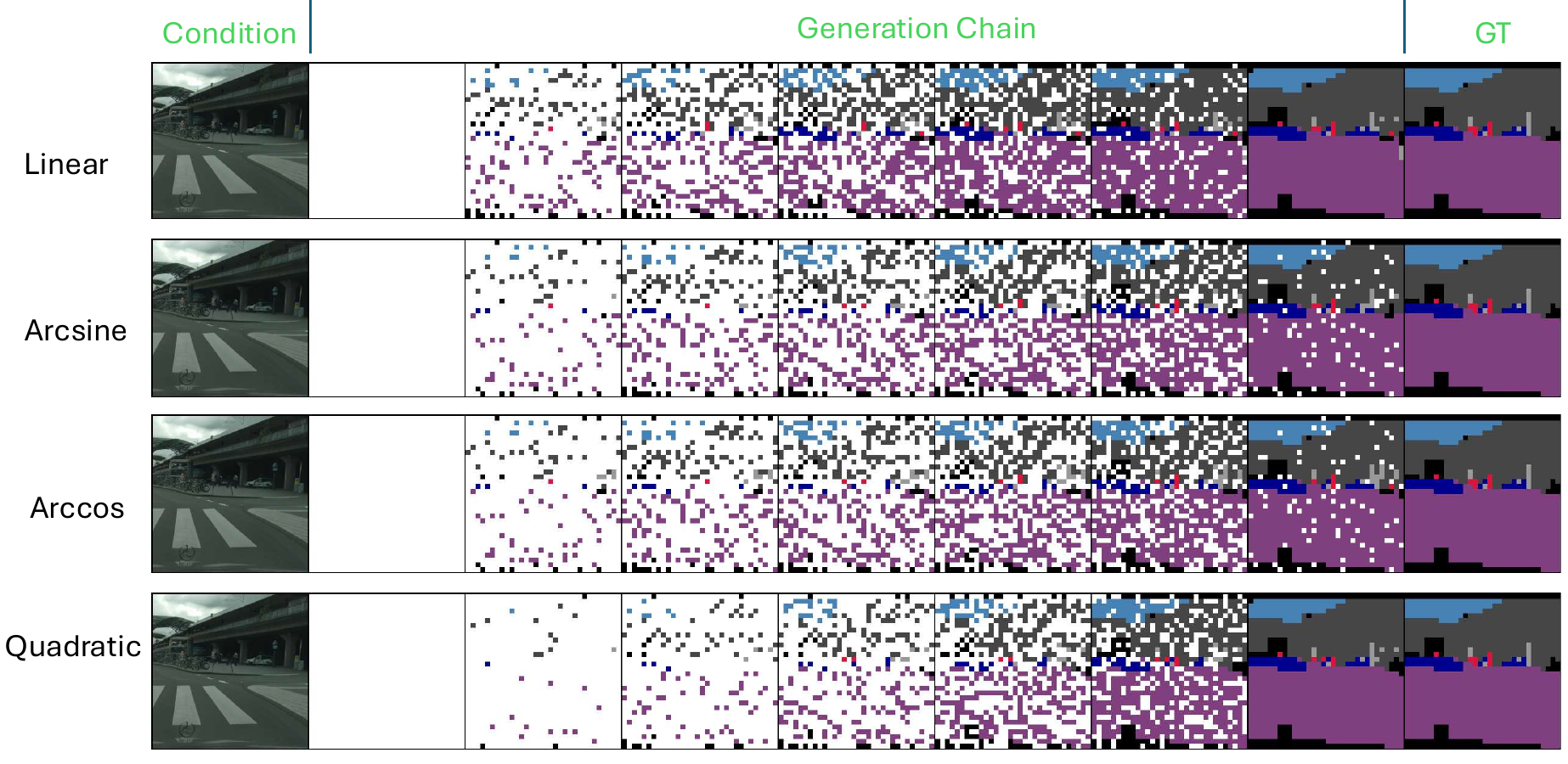}
    \caption{\textbf{Misalignment between schedulers. The progressive chain visualization of changing} when the sampling scheduler $\kappa(t)$ when trained with linear schedulers. We sampled with 50 steps and CFG scale=3.}
    \label{fig:argmax_save_scheduler}
\end{figure}

\begin{figure}
    \centering
    \includegraphics[width=0.79\linewidth]{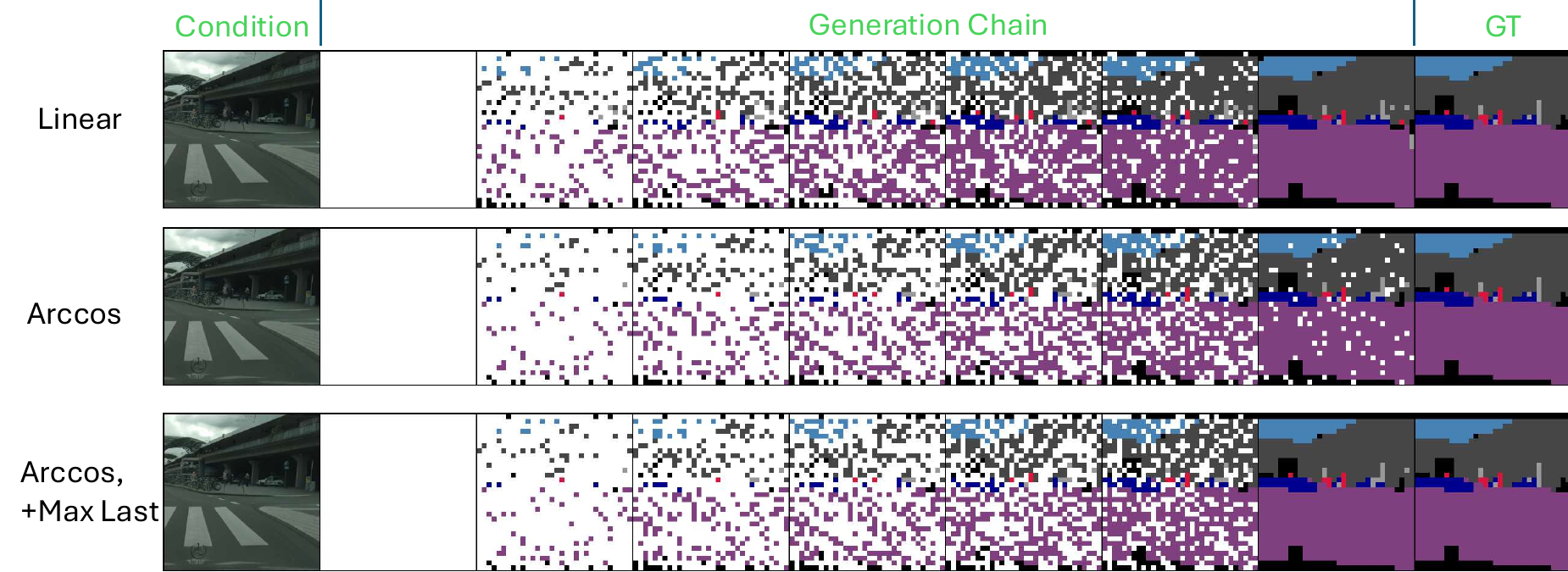}
    \caption{\textbf{\texttt{argmax} operation after logits can greatly alleviate the issue of those misalignments between training and sampling schedulers.} 
    } 
    \label{fig:enter-label}
\end{figure}

\section{Licences}

Datasets:
\begin{itemize}
     \item ImageNet~\cite{imagenet}: CC BY 2.0 license
    \item MS-COCO~\cite{coco}: Creative Commons Attribution 4.0 License
    \item FaceForensics: MIT license
\end{itemize}

Pretrained models:
\begin{itemize}
    \item Image autoencoder from Stable Diffusion~\cite{rombach2022high_latentdiffusion_ldm}: CreativeML Open RAIL-M License
\end{itemize}

\end{document}